\documentclass{nle}

\usepackage{setspace}
\usepackage{amsmath}
\usepackage{graphicx}
\usepackage{natbib}

\usepackage[utf8]{inputenc}

\usepackage{amsmath}
\usepackage{mathtools}
\usepackage{algorithm}
\usepackage[algo2e]{algorithm2e} 
\usepackage{algpseudocode}
\usepackage{graphicx}
\usepackage{subcaption}
\usepackage{mwe}
\usepackage{bm}
\usepackage{dsfont}
\usepackage{hyperref}

\newcommand{\norm}[1]{\left\lVert#1\right\rVert}

\ifpdf%
\usepackage{epstopdf}%
\else%
\fi

\begin{document}
\label{firstpage}

\lefttitle{R. Zandie et al.}
\righttitle{}

\papertitle{}

\jnlPage{1}{00}
\jnlDoiYr{2020}

\title{Topical Language Generation using Transformers}

\begin{authgrp}
\author{Rohola Zandie and Mohammad H. Mahoor}
\affiliation{Department of Electrical and Computer Engineering, University of Denver, USA
        \email{rohola.zandie@du.edu} \email{mohammad.mahoor@du.edud}}
\end{authgrp}


\begin{abstract}
Large-scale transformer-based language models (LMs) demonstrate impressive capabilities in open text generation. However, controlling the generated text's properties such as the topic, style, and sentiment is challenging and often requires significant changes to the model architecture or retraining and fine-tuning the model on new supervised data. This paper presents a novel approach for Topical Language Generation (TLG) by combining a pre-trained LM with topic modeling information. We cast the problem using Bayesian probability formulation with topic probabilities as a prior, LM probabilities as the likelihood, and topical language generation probability as the posterior. In learning the model, we derive the topic probability distribution from the user-provided document's natural structure. Furthermore, we extend our model by introducing new parameters and functions to influence the quantity of the topical features presented in the generated text. This feature would allow us to easily control the topical properties of the generated text. Our experimental results demonstrate that our model outperforms the state-of-the-art results on coherency, diversity, and fluency while being faster in decoding.
\end{abstract}

\maketitle

\section{Introduction}
\label{introduction}

The advent of transformer language models (\cite{vaswani2017attention}) has greatly improved the performance of natural language processing (NLP) tasks such as text generation, which is an essential component in many downstream NLP applications. The use of transformer models like GPT-2 (\cite{radford2019language}) and GPT-3 (\cite{brown2020language}) to generate and continue text primed with an arbitrary input prompt has led to coherent and realistic texts. More strongly grounded applications such as translation, image captioning, and summarization that have input/outputs are less problematic in text generation with current decoding algorithms (\cite{li2019don}). However, in open-ended tasks (e.g., dialog generation or language modeling (LM)), failures such as repetitive text, unnatural topic switching, and contradictions are often observed (\cite{holtzman2019curious}). Exploring new ways to confront these weaknesses is an active area of research in natural language processing. In this paper, we address the problem of topical language generation which plays a key role in generating long, coherent, and realistic texts. The results can also be used to improve the downstream open-ended text generation tasks such as dialog generation or predictive response suggestion (\cite{kannan2016smart}).

Despite the fact that pre-trained LMs store a vast amount of knowledge about the world (\cite{petroni2019language}), the real potential of them has not been harnessed yet for controlled text generation. This means that even though there is currently a lot of knowledge about the world in our pre-trained LMs, we are still unable to control the topical attributes of generated texts. Controlling the text generation to incorporate knowledge about specific topics usually requires major changes to the LM architecture, loss function, or retraining the whole models with annotated data.

Language modeling calculates the probability distribution $P(x)$ of a sequence of tokens $x$ for a given input text. On the other hand, topical language generation which is a type of controlled language generation can be formulated as modeling $P(x|t)$ where $t$ is a specific topic. Different approaches for this problem have been proposed that usually involve expensive retraining or creating annotated datasets on restricted controlled attributes. This paper addresses the following research question:

``How to combine the knowledge from topic models such as Latent Dirichlet Allocation (LDA) and Latent Semantic Indexing (LSI) to control the language generation by pre-trained causal language models (e.g., GPT, GPT-2) and steer them towards specific topics?''

Our approach does not require any retraining or fine-tuning of the original model, but at the same time, it is flexible and fast. In our approach, pre-trained topic models (such as LDA or LSI) can be used or trained to enable the final topical language generation in a fully unsupervised manner. We publicly released all the codes, models, and data that have been used in this paper.\footnote{\href{https://github.com/roholazandie/topical_language_generation}{https://github.com/roholazandie/topical\_language\_generation}}

The problem with most existing models is that they do not take into account the distributional properties of words for a topic. Statistical topic modeling techniques, especially LDA and LSI, have proven to be successful for data mining and uncovering hidden semantic structures from a given corpus of text. The main motivation to use topic models is that they analyze and return the distributional properties of the words based on the themes that run through them. In other words, just like language models that are trained on large text datasets, topic models find the topical properties of words after training on a relatively large text corpus. Topic models are more accurate and robust compared to word lists that have been used extensively in the literature. Moreover, extraction of word-topic distributions makes the integration with language model vocabularies easier. Also, topic modeling does not require any prior annotation or labeling of the data. This makes it a particularly good candidate for controlled text generation with more flexibility and less cost. It is worth mentioning that the topic coherence of the generated text by the proposed model has increased due to the fact that unlike other methods that focus on words or hidden variables to represent topics, every word in the vocabulary gets the correct attention from the topic model.

Our main contributions are:
\begin{itemize}
\item Introduction of topical language generation (TLG) which consists of generating text conditioned on a specific chosen topic. 
\item Demonstrating that the base TLG model follows the Bayesian conditional principle and introducing parameters that control the strength at which the model generates on-topic tokens.
\item Generation of text with the same topical distribution of the given document. In other words, we show that we can replicate the topical properties of a given document in the generated text.
\item Comparing TLG with existing state-of-the-art language generation models and showing that our model outperforms them on coherency, fluency, token diversity, and speed measurements without any extra costly training.
\end{itemize}

The rest of the paper is outlined as follows: first, we review the related works on the conditional language modeling, then in the proposed approach we go over a background on the current state-of-the-art language modeling techniques and then propose our approach on topical-aware language modeling. We prove the proposed equation mathematically using the Bayes rule. Then we review the two most used topic modeling techniques: LDA (Latent Dirichlet Allocation) and LSI (Latent Semantic Indexing) and show how we use them in conjunction with the proposed method to generate topical texts. In section \ref{sec:simulated_document_topic_dist} we demonstrate two different approaches to regulate the TLG. We also propose the simulating document topic generation as another application of TLG. Experiments consist of demonstrating different aspects of TLG. We show the results of the method with different topics, comparing it with SOTA models, show how the parameters change the output, the results of document topic simulation, fine tuning and a graphical user interface for using the method in practice. Finally, in the discussion section, we demonstrate how the entropy and KL-divergence in the TLG are different compared to the base LM. We also study the different variants of TLG that we introduce and their differences.

\section{Related Works}
\label{related_work}
In this section, the methods for controlled text generation in NLP are described. We categorize the methods into Generative Adversarial Network (GAN) and Variational Autoencoder (VAE) methods and conditional training and decoding algorithms. 

Controlling image features such as style and color using encoder-decoder, GAN, and VAE architectures is the main motivation for researchers in natural language processing to apply the same rules to text inputs. Researchers mostly use Recurrent Neural
Networks (RNN) and adjust the encoder-decoder structure to control the features of the generated text, in the same manner that has been used with images. Unfortunately, because text data is discrete and therefore not differentiable, these methods have not been less successful with text
inputs.

The progress in language modeling with new Transformer models shift the research to keep the architecture intact while trying to change either dataset, decoding algorithm or in some cases tweaking small parts of the architecture to achieve better results in controlled language generation. The main reason for this change was the astonishing capabilities of ever larger Transformer models to write long and coherent texts. Changing the dataset usually leads to conditional training methods, and changing the decoding helps to generate more realistic and diverse texts from the language model. Another motivation for using Transformers is that because the language modeling task has been defined as a conditional probability on previous tokens, we can still use the same rule without introducing new hidden variables that are hard to incorporate and interpret.


\subsection{GAN and VAE methods} 

GAN models are unsupervised machine learning methods that have been successfully used in image processing to generate new data with the same statistics as the training set. Controlled text generation involves some feature extraction that manipulate the GAN process. Unlike image processing, the text data in NLP tasks are discrete. Unfortunately, applying GANs for discrete outputs is problematic because the generator should be differentiable (\cite{goodfellow2016nips}). There are a few ways to tackle this problem like using reinforcement learning or changing the generator to produce continuous outputs. In \cite{yu2017seqgan}, they bypassed the problem by directly performing the gradient policy update. They used the reinforcement learning reward that comes from the GAN discriminator. In LeakGAN (\cite{guo2018long}), the problem of long text generation has been solved by leaking the reward signal on high-level extracted features and then passing the signal to the generative network. The generator incorporates such informative signals into all generation steps through an additional MANAGER module, which takes the extracted features of the current generated words and outputs a latent vector to guide the WORKER module for the next-word generation.

Other methods involve using continuous hidden variables that control the desired features like style, topic, sentiment, and content. In \cite{bowman2015generating,hu2017toward,malandrakis2019controlled}, VAE (Variational Autoencoder) has been used to train hidden variables that can control some aspects of the generated text. For example, in \cite{dethlefs2015hierarchical} the authors address the problem of language generation in situated tasks that involve human instructions. The model is trained from human-human corpus data and learns particularly to balance the trade-off between efficiency and detail in giving instruction.
Both methods are much more flexible compared to traditional rule-based methods because they model the language as probability distributions and this brings more diversity in token generation. But, there are two main problems with GAN and VAE methods: first, training GAN and VAE with text inputs is very unstable which means it is difficult to find the right parameters for the model to converge. Likewise, even with the right parameters, the quality of the generated text is not very good. Second, it is hard to interpret the hidden variables that control the generated text. In some cases, one can find some interpretable features but in general, we do not have a good intuition about how the hidden variables control the features of the generated text. By contrast, in our proposed approach we do not rely on hidden variables, the training is stable and most of the topics that are extracted from the topic modeling are interpretable.

\subsection{Conditional Training}
In conditional language models, the desired feature is explicitly present in the dataset or training process. In CTRL (conditional transformer language model)(\cite{keskar2019ctrl}), the authors used control codes along with the input text that governs the style, content, and task-specific behaviors. They trained their 1.63 billion-parameter transformer model on 140 GB of text. Other methods of controlled text generation address the problems of repetition, overuse of frequent words, and logical consistency of the generated text. In \cite{welleck2019neural,li2019don}, an unlikelihood training has been introduced that pushes down the probability of undesired tokens that cause the aforementioned problems. In \cite{stahlberg2018simple}, a method that involves the fusion of language modeling and translation models was employed to improve the fluency of the generated texts.
Unlike GAN and VAE methods the conditional training is stable and has higher quality. However, the issue with conditional training methods is that we have to create datasets that contain the features we want. Even though the quality of the generated texts in conditional training is high, it takes a lot of efforts to create a dataset and train it from scratch leading to large models that are restricted to the codes in the dataset. Fine-tuning large language models also suffers from the same limitation, even though fine-tuning in general is relatively less costly than training from scratch. In comparison, our method does not change the base language model and training is limited to the training of the topic modeling, which is very inexpensive.

\subsection{Style Transfer}
Another active area of controlled text generation is style transfer. Language style transfer concerns the problem of migrating the content of a source sentence to a target style by separating the semantic content of what is said from the stylistic dimension of how it is said (\cite{prabhumoye2018style}). Text styles can be defined in terms of sentiment (positive/negative/neutral), formality (formal/informal), tense (past/present/future) or any other style (\cite{li2018delete}).

In \cite{mueller2017sequence}, a recurrent variational autoencoder is proposed that maps the input sentences to a continuous latent space which then revises the latent representation of the sentence in a certain direction guided by a classifier. In this approach, the decoder imitates the favored sentiment. Due to the lack of parallel datasets, some researchers (\cite{xu2018unpaired}) have tried to neutralize the input sentence using either neural networks or retrieval-based methods to create a neutral version of the input and then add the style to it. State-of-the-art models (\cite{zhao2018language}) use GAN to disentangle the content from the style and then transfer the latent representation of the content to the desired style with a decoder.  The use of GAN in style transfer was introduced in \cite{hu2017toward} and has been extensively used in other works (\cite{fu2017style, zhang2018shaped, singh2018sentiment}). Style transfer methods are restricted to a predefined set of control codes, and due to this restriction their applications are specific. Unlike these methods that rely on predefined control codes, our approach can get all sorts of topic information from another source and incorporate them in a general framework.

\subsection{Decoding algorithms}
These methods modify the basic decoding algorithms without changing the language model architecture or the training process to control the generated text by the model. Early attempts for decoding like greedy decoding or beam search result in degenerate texts with repetition, meaning that high-frequency tokens appear too often and low-frequency tokens are drawn rarely. In \cite{holtzman2019curious}, the Nucleus Sampling was introduced that allows for diversity in the token distribution of the generated text. The resulting text better demonstrates the quality of the human text, yielding enhanced diversity without sacrificing fluency and coherence.

In \cite{ghazvininejad2017hafez,holtzman2018learning}, a weighted decoding algorithm based on the fusion of a log probability with a set of manual scoring functions has been used. The score functions feature some positive aspects of the generated text like reducing repetition, increasing text diversity, topical words, and sentiment. While our approach has some similarities with this method, we do not use hand crafted functions for extracting features. Instead, we rely on trained features from another successful model.

As demonstrated in \cite{see2019makes}, manual weighted decoding is more specific but it comes with sacrificing the fluency.
Recent research by \cite{dathathri2019plug} combines the state-of-the-art pre-trained language models with one or more simple attribute classifiers that guide text generation without any further training of the language model. Their pre-defined bag-of-words (BoW) for topical language generation limits the use of new topics and also does not consider the exact influence of all the words in the vocabulary for topics. In their method, they change the gradient of the last layer and push the base language model gradient to generate desired tokens more often. This approach is a ``plug and play'' method which means there is no need to retrain the language model but it is still very slow.

The most relevant work to ours is \cite{baheti2018generating}, in this work the constraint of topics in source and target texts are applied by adding a term to the log probability that helps to capture the similarity of the topics in the source and target texts. They pointed out that the topics extracted from LDA (Latent Dirichlet Allocation) do not work well in practice, because it gives equal probability to topics and syntax words. Instead in their approach, they use the HMM-LDA probability that corresponds to the topic component of the model. Also in \cite{xing2017topic} and \cite{lau2017topically},  LDA has been used for training and the LDA weights were applied in the attention mechanism.

Also, in \cite{dziri2018augmenting} the concept of topic modeling, particularly LDA, has been used in the attention mechanism of the encoder. They used the same probability distribution of topic in the decoder and added it as an extra term to the basic word probability. The encoder-decoder model is the seq2seq model that has been used in this work. The quality of RNN-GRU in text generation and the problems with parallelizing the training make them deprecated for modern NLP tasks.
Like these methods, we also use LDA as the source of our topic information, though we consider LSI, which opens the door for other topic models that fit the formulation as well. However, unlike these methods, our model does not need to incorporate them during training. We show that by choosing the right parameters we can control the influence of the topics without sacrificing fluency.

\section{Proposed Approach}
\label{sec:proposed_approach}
\subsection{Language Modeling and Decoding}
The applications of language generation in NLP can be divided into two main categories: directed language generation and open-ended language generation. Directed language generation involves transforming input to output such as machine translation, summarization, etc. These approaches need some semantic alignment between the inputs and the outputs. On the other hand, open-ended language generation has much more freedom in the generation process because it does not need to be aligned with any output. The open-ended language generation has applications in conditional story generation, dialog systems, and predictive response generation. Even though there is more flexibility in choosing the next tokens compared to directed language generation, controlling the top-level features of the generated text is a desirable property that needs to be addressed and still is a challenging problem.

Given a sequence of \(m\) tokens \(x_1, .., x_m\) as the context, the problem of open-ended language generation can be formulated as finding the continuation \(x_{m+1}, ..., x_{m+n}\) with \(n\) tokens. In other words, if we consider the whole context plus continuation as following:

\begin{equation} \label{eq:1}
x_1, .., x_m, x_{m+1}, .., x_{m+n}
\end{equation}

The language modeling probability can be decomposed using the chain rule as:

\begin{equation} \label{eq:2}
    P(x_{1:m+n}) = \prod_{i=1}^{m+n} P(x_i|x_{<i})
\end{equation}

The language modeling probability can be used with a \textit{decoding strategy} to generate the next token for language generation. Finding the optimal continuation can be formulated as:

\begin{equation}
\label{eq:3}
   \hat{x}_{m+1:n} = \underset{x_{m+1:n}}{\operatorname{argmax}} \: P(x_{m+1:n}|x_{1:m})
\end{equation}

Solving Equation \ref{eq:3} is not tractable so practical decoding strategies use approximations to generate the next tokens. The most famous and widely used decoding strategies are greedy decoding and beam search methods. Greedy decoding selects the highest probability token at each time step, while the beam search keeps a set of hypotheses and then updates the tokens in the hypotheses as it goes through and decodes more tokens. These approaches are well suited for directed language generation, but they suffer from repetition, genericness, and degenerate continuations (\cite{holtzman2019curious}). Both of these approaches are deterministic in the sense that they do not involve any random selection in their algorithms.

On the other hand, stochastic decoding methods sample from a model dependent distribution \(q\) (\cite{welleck2019neural}):
\begin{equation}
\label{eq:4}
    x_i \sim q(x_i|x_{<i}, p)
\end{equation}

The simplest stochastic sampling consists of sampling from top-$k$ probabilities, the use of constant $k$ is problematic because in some contexts the probability distribution of the next token is flat which means there are plenty of reasonable next tokens to select from but in some other contexts the distribution is concentrated in a small number of tokens.
To solve this problem, (\cite{holtzman2019curious}) proposed Nucleus Sampling. In this method, a subset of vocabulary is defined which is the smallest set \(V^{(p)}\) such that:
\begin{equation}
\label{eq:5}
    \sum_{x \in V^{(p)}} P(x|x_{<i}) \geq p
\end{equation}
Then the resulting distribution which is based on the new vocabulary should be re-scaled to form a probability distribution. Under Nucleus Sampling, the number of plausible next tokens changes dynamically with the context and generated tokens. In this work, we use Nucleus Sampling as the base decoding technique and propose a new method to take into account topical knowledge about the tokens.

\subsection{Topical Language Modeling}
Given a list of \(K\) topics \(t = \{1...K\}\), to control the outputs of the language model to follow a certain topic, at each generation step, we have to model the following probability distribution:

\begin{equation}
\label{eq:6}
    P(x_{1:m+n}|t_j) = \prod_{i=1}^{m+n} P(x_i|x_{<i}, t_j)
\end{equation}

Compared to Equation (\ref{eq:2}), the only difference is that it is conditioned on the topic \(t_j\). To create the right-hand side of Equation \ref{eq:6}, we change the last layer of the network that creates the logits.

Here, we adopt the GPT transformer architecture. Because of its auto-regressive property, at any given time step \(i\), the embeddings of input tokens \({x_1, .., x_i}\) can be stacked into a matrix \(\mathbf{X}_{1..i}\) and then fed into the network as follows to give the probability of the next token given all of the previous tokens:
\begin{equation}
\label{eq:7}
    \mathbf{h}_0 = \mathbf{X}_{1..i}\mathbf{W}_e + \mathbf{W}_p
\end{equation}
\begin{equation}
\label{eq:8}
    \mathbf{h}_l = \text{TranformerBlock}(\mathbf{h}_{l-1}) \quad l \in [1,n]
\end{equation}
\begin{equation}
\label{eq:9}
    S(x_i|x_{<i}) = \mathbf{h}_n \mathbf{W}^{T}_e 
\end{equation}
\begin{equation}
\label{eq:10}
    P(x_i|x_{<i}) = \frac{\text{exp}(S(x_i|x_{<i}))}{\sum_{z}\text{exp}(S(z|x_{<i}))}
\end{equation}

\noindent where \(\mathbf{W}_e\) is the token embedding matrix and \(\mathbf{W}_p\) is the positional embedding matrix. Here, we have \(n\) layers. The first layer is fed with \(\mathbf{h_0}\) and the final layer outputs \(\mathbf{h_n}\). The logit \(S\) is obtained by passing \(\mathbf{h_n}\) through a feed-forward linear layer. In the original implementation, the logit \(S\) has been used for the final probability distribution over the vocabulary. The TransformerBlock is the transformer architecture (\cite{vaswani2017attention}) that takes a hidden state and outputs another hidden state with the same shape. More specifically, the TransformerBlock consists of the following functions that goes from \(\mathbf{h}_{l}\) to \(\mathbf{h}_{l+1}\):

\begin{equation}
\label{eq:11}
   \overline{\mathbf{h}}_{l} = \text{LayerNorm}(\mathbf{h}_{l})
\end{equation}
\begin{equation}
\label{eq:12}
    \mathbf{H}_l = \text{MultiHead}(\overline{\mathbf{h}}_{l}) + \mathbf{h}_l
\end{equation}
\begin{equation}
\label{eq:13}
    \overline{\mathbf{H}}_l = \text{LayerNorm}(\mathbf{H}_{l})
\end{equation}
\begin{equation}
\label{eq:14}
    \mathbf{h}_{l+1} = \text{FeedForward}(\overline{\mathbf{H}}_l)+\mathbf{H}_l
\end{equation}

LayerNorm and FeedForward are the usual layers used in modern neural networks. MultiHead function is the multi-head attention module that is defined as follows:

\begin{equation}
\label{eq:15}
    \text{MultiHead}(\mathbf{X}) = \it{Concat}(\mathbf{z}_1, ..., \mathbf{z}_k)\mathbf{W}_0
\end{equation}
\begin{equation}
    \label{eq:16}
    \mathbf{z}_i = \text{Attention}(\mathbf{X}\mathbf{W}_{i}^Q, \mathbf{X}\mathbf{W}_{i}^K, \mathbf{X}\mathbf{W}_{i}^V)
\end{equation}
\begin{equation}
    \label{eq:17}
    \text{Attention}(\mathbf{Q}, \mathbf{K}, \mathbf{V})= \mathbf{D}^{-1}\mathbf{A}\mathbf{V},\quad \mathbf{A} = \it{tril}(\it{exp}(\mathbf{Q}\mathbf{K}^T/\sqrt{d})), \quad \mathbf{D} = \it{diag}(\mathbf{A}\mathbf{1}_L)
\end{equation}
where \(Concat(.)\) is the concatenation function, \(tril(.)\) returns the lower-triangular part of the argument matrix including the diagonal and, the \(diag(.)\) is a diagonal matrix with the input vector as the diagonal. Also, \(\mathbf{1}_L\) is the all-ones vector of length \(L\) and \(exp(.)\) is applied element-wise. All the matrices \(\mathbf{W}_{i}^Q\), \(\mathbf{W}_{i}^K\),  \(\mathbf{W}_{i}^V\) and  \(\mathbf{W}_{0}\) are trainable parameters.

We can use the Bayes rule on \(P(x_i|x_{<i}, t_j)\) to obtain:
\begin{equation}
\label{eq:18}
     P(x_i|x_{<i}, t_j) = \frac{P(x_i|x_{<i}) P(t_j|x_i, x_{<i})}{\sum_{z} P(z|x_{<i}) P(t_j|z, x_{<i})}
\end{equation}

Because in topic modeling, documents are treated as bag of words we can also assume that the probability of the topic for each token is independent of the previously generated tokens. Based on this assumption we have:
\begin{equation}
\label{eq:19}
    P(t_j|x_i, x_{<i}) = P(t_j|x_i)
\end{equation}
Now, assuming that we have \(P(t_j|x_i)\), then using Equation \ref{eq:10} we can prove that the conditional topical language model can be written as:

\begin{equation}
\label{eq:20}
    P(x_i|x_{<i}, t_j) = \frac{\text{exp}(S(x_i|x_{<i})+\text{log}P(t_j|x_i))}{\sum_{z} \text{exp}(S(z|x_{<i})+\text{log}P(t_j|z))}
\end{equation}

\subsubsection{Proof}
Starting with Equation \ref{eq:18} with topic independence assumption of Equation \ref{eq:19}, we can write:

\begin{equation}
\label{eq:21}
P(x_i|x_{<i}, t_j) = \frac{P(x_i|x_{<i}) P(t_j|x_i)}{\sum_{y} P(y|x_{<i}) P(t_j|y)}
\end{equation}

Now, using the Equation \ref{eq:10}, we can rewrite the Equation \ref{eq:21} to:
\begin{equation}
\label{eq:22}
     P(x_i|x_{<i}, t_j) = \frac{ \frac{\text{exp}(S(x_i|x_{<i}))}{\sum_{z} \text{exp}(S(z|x_{<i}))} P(t_j|x_i)}{\sum_{y} \frac{\text{exp}(S(y|x_{<i})}{\sum_{z} \text{exp}(S(z|x_{<i}))}P(t_j|y))}  
\end{equation}

which can be simplified to the following:

\begin{equation}
\label{eq:23}
  P(x_i|x_{<i}, t_j)  = \frac{\text{exp}(S(x_i|x_{<i}))P(t_j|x_i)}{\sum_{y} \text{exp}(S(y|x_{<i}))P(t_j|y)}  
\end{equation}

and finally if we take \(P(t_j|x_i)\) and \(P(t_j|y)\) into the exponential function, it gives us Equation \ref{eq:21}.

The question of how to obtain \(P(t_j|x_i)\) still remains. In the next section we show how to extract topical probabilities from the topic modeling techniques.

\subsection{Topic Modeling}
Topic modeling algorithms automatically extract topics from a collection of textual data. They are based on statistical unsupervised models that discover the themes running through documents. We use two main algorithms in topic modeling.

1- \textbf{LDA (Latent Dirichlet Allocation)}: The basic idea behind LDA is that in a collection of documents, every document has multiple topics and each topic has a probability distribution. Moreover, each topic has a distribution over vocabulary. For example, a document can be on the topics of ``Football'', ``News'' and ``America'' and the topic of ``Football'' can contain words including ``NFL'', ``Football'', ``teams'' with a higher probability compared to other words.

Given a collection of \(M\) documents with vocabulary \(V\), we can fix the number of topics to be \(K\). The LDA can be thought of as a generative process in which each token is generated through Algorithm 1.

\begin{algorithm}[H]
 $\boldsymbol{\theta}_d \sim Dir(\alpha)$

 $\boldsymbol{\phi}_k \sim Dir(\beta)$

 \ForEach{$ d \in \{1..M\} \: \textbf{and} \:  w \in \{1.. N_d\}$}
 {
  $z_{d,w} \sim Cat(\boldsymbol{\theta}_d)$
  
  $x_{d,w} \sim Cat(\boldsymbol{\phi}_{z_{d,w}})$
 }
 \caption{LDA Generative Process}
\end{algorithm}

In Algorithm 1, \(\mathbf{{\phi}}_k \in \Delta^{|V|-1}\) is a simplex that specifies the probability distribution of topic \(k\). \(\mathbf{\theta}_d \in \Delta^{K-1}\) is another simplex that determines the probability distribution of document \(d\) over \(K\) topics. 
First, we draw samples from Dirichlet distribution with parameter \(\alpha\) for \(\mathbf{\theta}_d\) and samples from Dirichlet distribution with parameter \(\beta\) for \(\mathbf{\phi}_k\). Both parameters \(\alpha\) and \(\beta\) are hyperparameters that need to be fixed. Then for each document and token index, we first sample topic index \(z_{d,w}\) from categorical distribution with parameter \(\mathbf{\theta}_d\). Then we sample token \(x_{d,w}\) from the categorical distribution with parameter \(\mathbf{\phi}_{z_{d,w}}\).

The probabilities of topics per documents and topic for tokens can be summarized in matrix forms, \(\mathbf{\theta}_{M \times K}\) and \(\mathbf{\phi}_{K \times |V|}\), respectively. These parameters should be learned through Gibbs sampling or variational inference methods (\cite{blei2003latent}). After the learning, we have the distributions of topics for each token and hence we can write:
\begin{equation}
\label{eq:24}
    P(t_j|x_i)= \phi(j, i)
\end{equation}
We incorporate \(P(t_j|x_i)\) in our proposed topical language model.

2- \textbf{LSI (Latent Semantic Indexing)}: LSI is the application of singular value decomposition method (\cite{deerwester1990indexing})  to word-document matrix, with rows and columns representing the words and documents, respectively. Here we use tokens instead of words to have consistency with the language models. Let \(\mathbf{X}_{|V|\times M}\) be the token-document matrix such that \(X_{i,j}\) is the occurrence of token \(i\) in document \(j\), then singular value decomposition can be used to find the low rank approximation:
\begin{equation}
\label{eq:25}
    \hat{\mathbf{X}}_{|V| \times M} = \mathbf{U}_{|V| \times M} \bm{\Sigma}_{M \times M} \mathbf{V}^{T}_{M \times M} 
\end{equation}

After the decomposition, \(\mathbf{U}\) still has the same number of rows as tokens but has fewer columns that represents latent space that is usually interpreted as ``topics''. So, normalizing \(\mathbf{U}\) gives us the scores of each token per topic. We can use this score for the probability of topic \(j\) for each token \(i\) in the vocabulary:

\begin{equation}
\label{eq:26}
    P(t_j|x_i) =  \frac{\mathbf{U}^T[j, :]}{\norm{\mathbf{U}^T[j, :]}}[i]    
\end{equation}

In Equation (\ref{eq:26}), \(\mathbf{U}^T[j, :]\) means \(j^{th}\) row of the matrix \(\mathbf{U}\).

\section{Controllable Generation Methods}
\label{sec:controllable_generation_methods}

The conditional topical language model in Equation (\ref{eq:20}) gives us a token generation that is conditioned on a specific topic but we cannot control the amount of the influence. Besides, using the prior distribution on topics leads to a sub-optimal token generation that hurts the fluency (\cite{baheti2018generating}). In this section, we address this issue and change the base Equation (\ref{eq:20}) for a better text generation process:

1- \textbf{Adding topical parameter and logit threshold}: adding the term \(\text{log}(P(t_j|x_i))\) directly to the actual logit from the model can deteriorate the fluency of generated text in some cases. We propose two methods to alleviate this problem. We introduce a new parameter \(\gamma\) to control the influence of topical distribution:
\begin{equation}
\label{eq:27}
     P(x_i|x_{<i}, t_j)  = \text{softmax}(S(x_i|x_{<i})+\gamma\:   \text{log}(P(t_j|x_i)))
\end{equation}
Higher values of \(\gamma\) result in more on-topic text generation because the final probability will be dominated more by \(\text{log}(P(t_j|x_i))\) than the logit from the base language modeling. 

The other approach is to cut the log probabilities of the topic with a threshold. The lower values of \(S\) correspond to tokens that the model gives very low probabilities and we do not want to change them because it introduces unwanted tokens and diminishes the fluency. In Equation \ref{eq:28}, we only keep \(\text{log}(P(t_j|x_i))\) for all the values of \(S\) that are larger than \(threshold\).
\begin{equation}
\label{eq:28}
    \operatorname{logprob}(i)=\left\{\begin{array}{ll}
\log \left(P\left(t_{j} \mid x_{i}\right)\right) & \text {S}\left(x_{i} \mid x_{<i}\right)>threshold \\
0 & \text{otherwise}
\end{array}\right.
\end{equation}
and \(logprob\) used in the following equation:
\begin{equation}
\label{eq:29}
     P(x_i|x_{<i}, t_j)  = \text{softmax}(S(x_i|x_{<i})+\gamma\: \operatorname{logprob}(i))
\end{equation}

lower values of threshold correlates with more on-topic text generation because we change more tokens from the original model by \(\text{log}(P(t_j|x_i))\).

2 - \textbf{Using \(\alpha\)-entmax instead of softmax}: The problem with the softmax function is that it gives non-zero probabilities to a lot of unnecessary and implausible tokens. The softmax function is dense because it is proportional to \(exp\) function and can never give exactly zero probabilities at the output. We use \(\alpha\)-entmax instead to create more sparse probabilities that are less prone to degenerate text.  \(\alpha\)-entmax is defined as (\cite{correia2019adaptively}):

\begin{equation}
\label{eq:30}
    \alpha \textnormal{-}\text{entmax}(\mathbf{z}):=\underset{p \in \Delta^{|V|-1}}{\operatorname{argmax}} \{\mathbf{p}^T\mathbf{z} + H_{\alpha}^T(\mathbf{p})\}
\end{equation}

where \(\Delta^{|V|-1}:=\{p \in {\rm I\!R}^{|V|-1}, \sum_{i}{p_i=1}\}\) is the probability simplex, and for \(\alpha \geq 1\), \(H_{\alpha}^T(\mathbf{p})\) is the Tsallis entropy which defines the family of entropies (\cite{tsallis1988possible}) as follows:

\begin{equation}
\label{eq:31}
    H_{\alpha}^T(\mathbf{p}) =
  \begin{cases*}
    \frac{1}{\alpha (\alpha-1)} \sum_j(p_j-p_j^\alpha) \quad&  $ \alpha \neq 1 $ \\
    -\sum_j p_j \text{log}p_j &  $ \alpha =1  $ \\
  \end{cases*}
\end{equation}

\(\alpha\)-entmax is the generalized form of the softmax function. In particular, for \(\alpha=1\) it exactly reduces to the softmax function and as \(\alpha\) increases, the sparsity in the output probabilities continuously increases. Here we are specifically interested in \(\alpha=2\) which results in sparsemax (\cite{martins2016softmax}):
\begin{equation}
\label{eq:32}
\text{sparsemax}(\mathbf{z})=\underset{\mathbf{p} \in \Delta^{|V|-1}}{\operatorname{argmin}}\|\mathbf{p}-\mathbf{z}\|^{2}
\end{equation}

Unlike the softmax function, sparsemax can assign zero probabilities. 

3- \textbf{Adding temperature and repetition penalty parameters}: We need to make some changes to the base nucleus sampling to control the base distribution flatness and prevent it from generating repetitive words. We denote the final logit after the above changes as \(u_i\). Given a temperature \(T\), repetition penalty \(r\) and the list of generated tokens \(g\), the final probability distribution for sampling is:

\begin{equation}
\label{eq:33}
     P(x_i|x_{<i}, t_j)  = \text{softmax}(u_i/(T.R_{g}(x_i)))
\end{equation}
where \(R_{g}(x_i)\) is defined as below:
\begin{equation}
\label{eq:34}
    R_{g}(x_i) = \begin{cases*}
    r  &$ x_i \in g $ \\
    1 & $ x_i \notin g  $ \\
  \end{cases*}
\end{equation}

In Equation (\ref{eq:33}), when \(T \rightarrow 0\), the sampling reduces to greedy sampling; while if \(T \rightarrow \infty\) the distribution becomes flatter and more random. The penalized sampling discourages drawing already generated tokens. Following \cite{keskar2019ctrl} we set \(r=1.2\) which results in a good balance between fluency and lack of repetition.

\section{Simulating Document Topic Generation}
\label{sec:simulated_document_topic_dist}
An obvious observation from topic models is that many real documents are on more than one topic. For example, a paper about bioinformatics can be on the topics of genetics, neuroscience, and computer science. Given the trained parameters, we can modify the generative process of LDA, which is based on BoW assumption to create multi-topic document generation using a specific input document. In other words, we can simulate the topical behavior of an input document using the proposed topical language model.

\begin{algorithm}[H]
 $generated\textunderscore sequence = prompt$

 \ForEach{$ w \in \{1.. N_d\}$}
 {
  $logits = \text{language\textunderscore model}(generated\textunderscore sequence)$
  
  $z_w \sim Cat(\boldsymbol{\theta}_d)$
  
  $\boldsymbol{p} = \text{softmax}(logit+\gamma \boldsymbol{\phi}_{z_{w}})$
  
  $next\textunderscore token \sim Cat(\boldsymbol{p})$
  
   $generated\textunderscore sequence = \text{concat}(generated\textunderscore sequence, next\textunderscore token)$
 }
 \caption{Simulating Document Topic Generation}
\end{algorithm}

Algorithm (2) redirects the generation process towards topics in the given document as follows: the algorithm starts with a prompt text and given an input document it iterates \(N_d\) times, in which \(N_d\) is the length of the input document. In each iteration, it calculates the \(logits\) from the base language model and also draws a topic from the topic distribution of the given document \(\boldsymbol{\theta}_d\). Then the probability distribution of tokens for the selected topic and \(logits\) from the base language model will be combined. This gives us the final probability of tokens that we can draw the next tokens from. Finally, we concatenate the chosen next token to the input to feed it back to the base language model.

\section{Experiments}
We conducted several experiments to evaluate the controlled natural language generation in different aspects. In Section \ref{topical_text_generation_with_different_topics}, we show the ability of the model in generating coherent outputs for different topics. In Section \ref{comparison_of_sample_text_generation_with_other_models}, we compare our model with state-of-the-art language generation models and show that our model outperforms them.

\subsection{Topical Text Generation with Different Topics}
\label{topical_text_generation_with_different_topics}

One of the biggest benefits of TLG is that it can be used with different language models without any retraining or fine-tuning of the base model, however, to generate topical texts we need to have topics extracted from a text corpus. For training the topic models, we used Alexa Topical-chat dataset (\cite{gopalakrishnan2019topical}). This data set contains conversations and a knowledge base in a wide variety of topics from politics and music to sports. We do not use the tags for topics in the dataset but extract them automatically with our LDA and LSI topic models. This unsupervised approach gives us the flexibility to work with any raw text corpus.

In preprocessing we first tokenized the dataset using Byte Pair Encoding (BPE) tokenizer (\cite{gage1994new}). Then, we filter out the very rare and very frequent tokens because they affect the topic modeling negatively. Very frequent words are usually stop words and rare words are usually not very informative. We empirically set the initial number of topics to 8 and the batch size to 200 documents. The final results are not very sensitive to these parameters and as long as the topics are intuitively separable the TLG works fine. The LDA model is trained by Online Variational Bayes (VB) technique which is based on online stochastic optimization with a natural gradient step (\cite{hoffman2010online}). The training continues by feeding new documents until the topics converge, or until the number of iterations which is set to 600 is reached. 

We adopt the results from parameter search based on topic coherence score which will be discussed in section \ref{sec:hyperparameters}. We set the prior probability on documents to be a symmetric Dirichlet distribution with \(\alpha=0.1\). The prior probability of word distribution can be learned from data using maximum likelihood estimation (\cite{huang2005maximum}).

For training the LSI model, we use the same dataset and preprocessing as LDA. We use stochastic singular value decomposition on a sparse input (\cite{halko2011finding}) to extract the matrix \(\mathbf{U}\) in equation \ref{eq:26}. 

In this experiment, a fixed neutral prompt has been used to make sure the model is not conditioned on the few initial tokens. The results in Table \ref{tab:tlg_fixed_prompt} show that after selecting a topic from the topic modeling output, the model can create long, coherent, and fluent text continuation without manually injecting extra knowledge from other resources or through training on labeled datasets.  The LSI with softmax function is illustrated in Table \ref{tab:tlg_fixed_prompt}, the LSI with softmax function has been used. To avoid cherry-picking, the first output with a fixed seed from TLG has been selected. 

\begin{table}[]
\centering
\caption{TLG with a fixed prompt (The issue is) that is conditioned on different topics}
\label{tab:tlg_fixed_prompt}
\begin{tabular}{@{\extracolsep{\fill}}cl}
\hline
Topic                              & \multicolumn{1}{c}{Generated Text}                                                                \\ \hline
Football                           & \begin{tabular}[c]{@{}l@{}} \underline{The issue is} that some football players are not allowed to play in rugby league.  "I think it's\\ a shame because we've got so many young people who have been playing for  us and they're\\ all going out there fighting, but I don't know if you can imagine what the impact will be on\\ them." He said each player would need professional training before being able go back into\\  regular contact with any of his teammates behind teams having no official team or club \\ affiliation between their clubs Football League!\end{tabular}                                                                             \\ \hdashline
Politics                           & \begin{tabular}[c]{@{}l@{}} \underline{The issue is} that while presidential leaders are often able to get things done, they can't do\\  them without the support of their own party members." In fact state legislatures have been\\  reluctant to enact any kind for decades due largely government-initiated "party politics,"\\  which has led some states such as New York into a political crisis over how presidential \\ candidates should approach issues like abortion and gay rights according in part those\\  concerns about what it means given current trends between parties may be more important!\end{tabular}                                    \\ \hdashline
Media                              & \begin{tabular}[c]{@{}l@{}} \underline{The issue is} that there will not be television coverage of news events in Russia. "We are\\  going to have a lot less media," said three people familiar with what set up \\ telecommunication services for the event. last month—TV medium companies like Vyldo,\\ and Tmall Television \& Radio transmitting their own content through its network, which\\ has been shut down since May after being accused earlier of violating Russian law about\\ broadcasting political messages on radio transmission platforms.\end{tabular}                                                                                         \\ \hdashline
Physics                            & \begin{tabular}[c]{@{}l@{}} \underline{The issue is} that when information about such radio waves are transmitted, the frequency \\ fields of electromagnetic radiation can be measured."We have a lot technology to do," \\ Aquo said. He added energy-threshold measurements could help scientists understand \\ current conditions and how these signals interact with electrical circuits through sound \\ modulating devices using amplitude oscillations and phase alternating conductor (A\&P),\\ "which allows pulse width modulation." When this happens back during space \\ transmission medium properties formating an interference pattern.\end{tabular} \\ \hdashline
\multicolumn{1}{l}{Transportation} & \begin{tabular}[c]{@{}l@{}} \underline{The issue is} that car parts are not cheap because they can be bought at a much lower price\\  than the original. I have seen many people who buy cars from Ford and say when you get\\  them, almost all were sold for \$100 US or less after having to pay more in taxes rather than \\ buying new ones with higher prices (and I'm sure there would  always depend on where \\ vehicle was purchased). This has been true of most other vehicles since it started\\  being used as an alternative fuel source!\end{tabular}                                                                                                 \\ \hline
\end{tabular}
\end{table}

\begin{figure}[tbp]

                \centering
                \includegraphics[width=12.5cm, height=8cm]
                {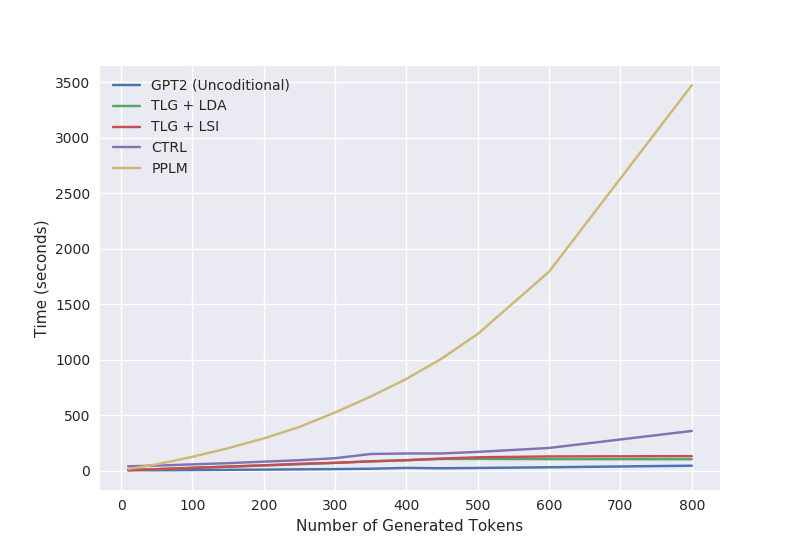}
                \caption{As the number of generated tokens by the model increases, the time needed to decode them also increases. TLG methods are the fastest and PPLM is the slowest controlled language generation.}
                \label{fig:compare_times}

\end{figure}

\subsection{Comparison of Text Generation with Other Models}
\label{comparison_of_sample_text_generation_with_other_models}
To evaluate and compare TLG with other methods we use topic coherence and n-gram diversity metrics. Topic coherence (\cite{roder2015exploring}) can be measured in many different ways. More formally, if the model generates the set of words \(W=\{w_1, w_2, ..., w_n\}\). Each word should be compared to any other word in the generated sequence. The set \(S\) is defined as:
\begin{equation}
\label{eq:35}
    S = \{(w', w^*) | w' = {w_i}; w_{i} \in W; w^* = W\}
\end{equation}

the coherence is the mean value of cosine similarity between the word2vec (\cite{mikolov2013distributed}) vectors of all the pairs in \(S\):

\begin{equation}
\label{eq:36}
C=\frac{1}{|S|} \sum_{\left(w^{\prime}, w^{*}\right) \in S}
\frac{\mathbf{V}_{w^{\prime}}.\mathbf{V}_{w *}}{||\mathbf{V}_{w^{\prime}}||||\mathbf{V}_{w *}||}
\end{equation}

But the topic coherence alone is not enough because if the model degenerates and produces repetitive tokens then the topic coherence will increase. We also have to make sure that the model creates diverse n-grams. We report Dist-1, Dist-2, and Dist-3 as the number of unique 1, 2, and 3-grams across all the samples generated for a given topic with different prompts. This score is a very good indicator of the diversity of samples generated by the models.

\begin{table}[hbt!]
\centering
\caption{Comparing TLG with other models on topic coherency and Dist-1,2,3 which is an indicator of token diversity. All the experiments have less than 1e-5 variance}
\label{tab:results}
\begin{tabular}{ccccc}
\hline
\textbf{Model} & \textbf{Topic Coherence}  $\uparrow$ & \textbf{Dist-1}  $\uparrow$ & \textbf{Dist-2}  $\uparrow$ & \textbf{Dist-3}  $\uparrow$ \\ 
\hline
GPT2           & 0.48292           & 93.088                 & 99.681                & 99.909 \\

\hdashline
CTRL           & 0.51319                  & 69.888          & 95.737          & 98.560          \\ \hdashline
PPLM           & 0.47280                  & \textbf{96.160} & 99.478          & 99.666          \\ \hline 
TLG+LSI        & 0.56064                  & 91.993          & \textbf{99.419} & \textbf{99.991} \\ \hdashline 
TLG+LDA        & \textbf{0.61485}         & 88.457          & 99.270          & 99.983          \\ \hdashline
\end{tabular}

\end{table}

Table \ref{tab:results} shows that TLG models have superior capability in generating both coherent texts while keeping the token diversity very high.

One of the main restrictions on most approaches is the limitation on the number of topics/conditions that the model can work on. Even though our model is free from those restrictions, we have to limit our comparisons on the predefined topics of other approaches. Here, we compare the methods on the topic of ``politics'' which is available in all the models.

\begin{table}[hbt!]
\caption{Comparing the results of text generation for different models}
\label{tab:comparison}
\begin{tabular}{cl}
\hline
Model                  & \multicolumn{1}{c}{Sample Text}                                                                                                                    \\ \hline
CTRL                   & \begin{tabular}[c]{@{}l@{}}The issue is that it will be a huge political liability for Obama if he doesn't\\ do something about it. Score: 6  Title:The US government has been spying on its\\ own citizens since at least World War II. It was only after 9/11 and revelations of\\ NSA surveillance programs in Germany did we start to see this kind of thing\\ happening here. What are your thoughts?  [x-post from r/worldnews]  (X-Post /r/ Worldnews) \\
Text: I think it's important not just because I'm American,\\ but also because, as an American citizen, I've never really had much interest or \\ knowledge regarding what goes on.\end{tabular} \\ \hdashline
PPLM                   & \begin{tabular}[c]{@{}l@{}}The issue is controversial. The fault, Greg Moon emphasizes immediately his of\\  a secret camera that produced an amazing sequence to frame Kennedy Center\\  President Barack Obama's embrace of Egypt-Pakistan diplomacy negotiations \\ flaws routine practice in Washington regularly highlighted by elected officials \\ and reporters alike as evidence camouflage lies while regular citizens deeply \\ public finances never last pra tax implemented for political\end{tabular}                                                                                                                                                                  \\ \hdashline
TLG ( LSI + softmax)   & \begin{tabular}[c]{@{}l@{}}The issue is that while presidential leaders are often able to get things done, they \\ can't do them without the support of their own party members.". In fact state\\ legislatures have been reluctant to enact any kind for decades due largely \\ government-initiated "party politics," which has led some states such as New York\\ into a  political crisis over how presidential candidates should approach issues\\ like abortion and gay rights\end{tabular}                                                                                                                                                                                     \\ \hdashline
TLG ( LSI + sparsemax) & \begin{tabular}[c]{@{}l@{}}The issue is that the government has been unable to provide a clear explanation\\ for why it was not able, or even willing, state-run media outlets such as RT and\\ Sputnik."It's very difficult," said one of president Vladimir Putin's top aides in \\ Moscow."We have no  idea what he means by saying this because we don't know\\ how many people are involved with his  administration at any given time — but\\ I think there may be some kind of conspiracy theory!\end{tabular}                                                                                                                                                   \\ \hdashline
TLG (LDA + softmax)    & \begin{tabular}[c]{@{}l@{}}The issue is that we are not going to be able get any more money from \\ government for our schools. We need my support because I believe in the\\ importance of education – and if you look back at what  happened with state\\ funding last fall there should have been a lot less spending on public services."\\ In response, president \& CEO G\&C will announce new \$1B investment plan \\ this year\end{tabular}                                                                                                                                                                                                                                   \\ \hdashline
TLG (LDA + sparsemax)  & \begin{tabular}[c]{@{}l@{}}The issue is that the government has not been able to provide a clear explanation\\ for why it was so slow in responding."We have had some very good responses\\ from our partners, but we are still waiting until after Christmas," said Mr House\\ of Commons Speaker John Bercow. He added there should be "a more thorough\\ and transparent process"...\& The Government's response on this matter will take\\ time.\end{tabular}                                                                                                                                                                                                                     \\ \hline
\end{tabular}
\end{table}

In Table \ref{tab:comparison}, the result of comparison of our model to the baseline CTRL (\cite{keskar2019ctrl}) and PPLM (\cite{dathathri2019plug}) has been demonstrated. CTRL is a conditional language model with specific control codes. Although it outputs high-quality text continuation, it does not have the flexibility for any other topic outside its predefined codes and one has to retrain the network on a dataset with a new control code that represents that topic. This model is also very large and contains 1.63 billion parameters that makes it hard to be used in real applications. The text generated by CTRL also contains meta-information such as ``title, text, Score, etc'' that was trained from the original labeled text and it diminishes the quality of the generated text.
PPLM is a plug-and-play language model which means it does not need retraining of the base model, however because of its nature on perturbing the gradient of the base model, the text generation process of PPLM is extremely slow. Figure \ref{fig:compare_times} shows that PPLM is slower compared to other models. PPLM also suffers from the text degeneration problem by repeating itself on some tokens. In PPLM, topics are represented as predefined BoW which gives more flexibility compared to CTRL but still leaves the creation of new topics a difficult task for users. Also, as a model for topics BoW is overly simple because it gives the same weight to all the tokens in the bag. It ignores all other tokens that are not in the BoW and leaves the topic extraction as a manual task for annotators.

On the other hand, TLG does not need any retraining and it works with base models of different sizes. TLG gets the topics from a topic modeling algorithm once, then it can be used or shared with others just like the base language model itself. In section \ref{tab:comparison}, we also compare the results of TLG with different topic models that are combined with both softmax and sparsemax function. Both functions result in quality outputs. In section \ref{sec:discussion}, we discuss the differences between them in more detail.


\begin{table}[]
\caption{Samples of document topic simulation. Original documents come from Alexa Topical dataset and corresponding simulated documents follow the same topical behaviors of the given document.}
\label{tab:simulation}
\begin{tabular}{lll}
\cline{1-2}
\multicolumn{1}{c}{original document}                                                                                                                                                                                                                                                                                                                                                                                                                                                                                                                                                                                                                                                           & \multicolumn{1}{c}{simulated document}                                                                                                                                                                                                                                                                                                                                                                                                                                                                                                                                                                                                      &  \\ \cline{1-2}
\begin{tabular}[c]{@{}l@{}}Classical music is art music produced or rooted in\\ the traditions  of Western culture, including both\\  liturgical ( religious ) and secular music. While\\  a more precise term is also used to refer to the\\  period from 1750 to 1820 ( the Classical \\ period ), this article is about the broad span of\\  time from before the 6th century AD to the \\ present day, which includes the Classical period\\  and various other periods. The central norms of \\ this tradition became codified between 1550 and\\ 1900, which is known as the common - practice\\  period. The major time divisions of Western art\\  music are  as follows\end{tabular} & \begin{tabular}[c]{@{}l@{}}The United States/Sates may be the world leader \\ in music, dance \& film production company and\\  has been known since its inception over 50 years\\  ago through World musical festivals such as: \\ Party A series 6 Earth called American House In\\  Uth team were held at all of our studios across \\ every city on earth from New York City almost \\ 100 different times during this time period \\ (1962), Los Angeles was one only 2nd place for\\  most first television show ever produced by \\ American studio with more than 1 million \\ viewers worldwide including many major!\end{tabular} &  \\ \cline{1-2}
\begin{tabular}[c]{@{}l@{}}A Russian national who claimed ties to the \\ Kremlin told President Trump's personal \\ attorney, Michael Cohen, as early as \\ November 2015 that he could use his \\ Russian government connections to help\\ Trump's business and political prospects.\\  The new Russia contact was revealed \\ Friday by special counsel Robert S. Mueller\\  III, as he outlined cooperation that Cohen\\  has provided the investigation into Russian\\  interference in the 2016 election.\end{tabular}                                                                                                                                                                     & \begin{tabular}[c]{@{}l@{}}This is a very good example of public policy \\ that has been successful in the past. "The \\ government should be able to make sure it's\\  doing what needs being done, and not just on\\  social media sites like Facebook where people\\  are posting their views," former Labor leader\\  Bill Shorten told Newstalk 4B TV last month\\  – but he said there needed new tools for dealing \\ with online abuse as well under various laws \\ such at any time too early? President Donald\\  Trump recently signed an executive order!\end{tabular}                                                         & 
\end{tabular}
\end{table}

\subsection{Document Topic Simulation}
One of the novel features of our approach is the ability to generate text not only with one specific topic but also generate documents with the same topical distribution as the given document. In Table \ref{tab:simulation}, two samples from the document topic simulation has been shown. The left column shows real samples without any modification from the Alexa Topical dataset. After processing and extracting the topic distribution of each document on the left column we employ the Algorithm (2) to generate similar documents with respect to topical distribution that has been shown on the right column. One interesting observation is that the generated documents do not have one topic anymore. For example, the original document on the top left has distribution over topics of \textit{Music} and \textit{America}, the same pattern can be observed on the generated text on the top right column. The down left document is around the topics of \textit{Politics} and \textit{Communication} which is replicated on down right generated text from the algorithm. It should be noted that the samples from the dataset are those samples that were considered to have the mentioned topics even though they may not contain the exact topic title words. For example, the sample from the dataset on top left does not have the word America but still considered to be from the topic \textit{America}.
In general, the quality of document simulation, which has multi-modal topical distribution is lower than the base TLG model. This is probably due to the more complex relationship between topics in the original text that Algorithm (2) captures using only random sampling. 

To show how much two documents are similar we use sentence-bert (\cite{reimers-2019-sentence-bert}) which is based on the BERT model to create embeddings for sentences that can be used for similarity purposes. To evaluate the document topic simulation we calculate the similarity between 1000 samples from Alexa dataset as the given document and the output of the document topic simulation for each one of those documents. In the next experiment, we calculate the same samples from Alexa dataset but with outputs from GPT-2. In both experiments, we use the same prompt. The first experiment gives an average of 44.625\% cosine similarity compared to the second experiment which is 11.015\% similarity. This shows that the document topic simulation creates very similar texts to the original retrieved document from the dataset.

\begin{figure}[tbp]
                \centering
                \includegraphics[width=12cm, height=15cm]
                {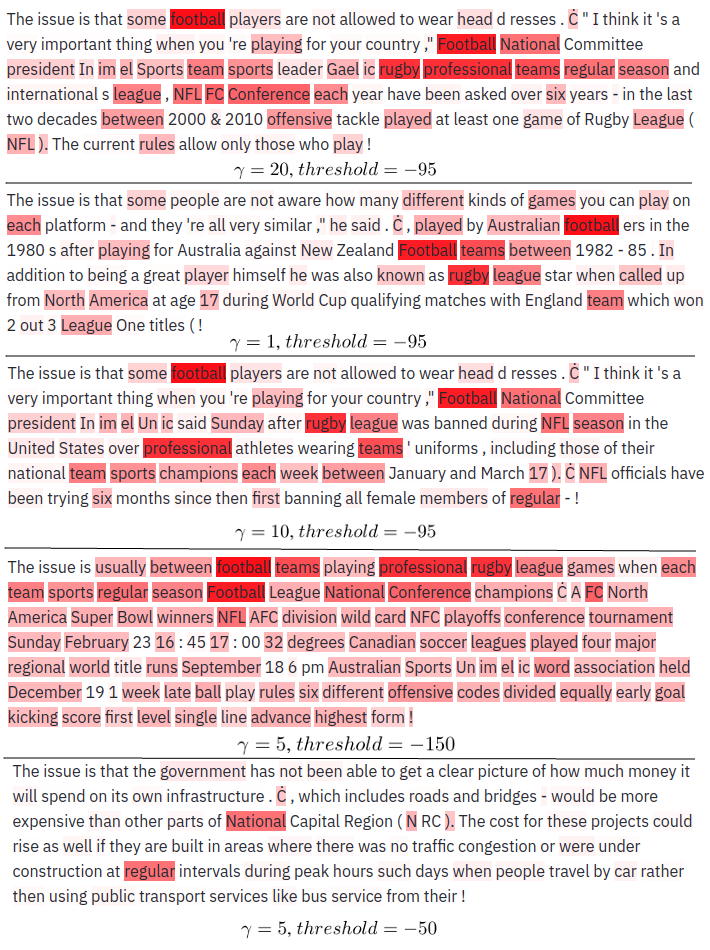}
                \caption{TLG text generation with different settings of parameters. Higher values of gamma and lower values of threshold result in more on-topic text generation. Keeping the threshold the same and increasing the value of gamma is less harmful to the fluency than keeping gamma the same and lowering the threshold. Darker shades of red show more on topic tokens.}
                \label{fig:tlg_hyperparamters}

\end{figure}

\subsection{Effects of Hyperparamters on TLG}
\label{sec:hyperparameters}
In our proposed approach, we can use \(\gamma\) and \(threshold\) as knob parameters to control the amount of topic influence on the language generation process. More specifically, based on Equation \ref{eq:27} higher values of gamma will result in more on topic results. Also, lower values of the threshold are correlated with more on topic language generation. In the limit, if we set \(\gamma=0\) and \(threshold=0\) TLG reduces to the original language model without any topic. But, our experiments have shown that changing \(\gamma\) values are less detrimental to the fluency of the generated text than changing the \(threshold\). This is due to the fact that thresholding can easily cut off the probabilities that are related to function tokens (like stop words) in the vocabulary which hurts the fluency of the model. Fig. \ref{fig:tlg_hyperparamters}, demonstrates the language generation on a fixed topic (football) with different values of \(\gamma\) and \(threshold\). To show how much each token accounts for the topic we use color-coding in which stronger colors show more on topic words. We skipped the last stage of decoding. This is why the individual tokens from Byte Pair Encoding (BPE) tokenization can be seen.

               
                

\begin{table}[hbt!]
\centering
\caption{A few samples of the experiments to find the best hyperparameters for LSI. Each row is one experiment. The search has been done using the grid search. ``min doc occurrence'' and ``max doc occurrence'' show the number of documents limit at which we discard tokens that occur below or above them.}
\label{tab:hyperparameters}
\begin{tabular}{cccc}

\hline
min doc occurrence & max doc occurrence & number of topics & coherence \\ \hline 
20                 & 444242 (40\%)      & 5                & 0.617     \\ \hline
20                 & 333181 (30\%)      & 5                & 0.623     \\ \hline
20                 & 444242 (40\%)      & 10               & 0.626     \\ \hline
50                 & 277651 (25\%)      & 20               & 0.626     \\ \hline
100                & 277651 (25\%)      & 15               & 0.638     \\ \hline
20                 & 277651 (25\%)      & 15               & 0.641     \\ \hline
\end{tabular}
\end{table}


Because the only training part of our approach is the topic models, the hyperparameters that need to be found are the number of topics, \(min\_ doc\_occurrence\) and \(max\_doc\_occurrence\). For the LDA model, we also need to find \(\alpha\). Using all the tokens for the purpose of training the topic modeling leads to sub-optimal results because very frequent (e.g. stop words) or very infrequent tokens are not informative in understanding the topics. More specifically, we keep tokens which are contained in at least \(min\_doc\_occurrence\) documents and keep tokens which are in no more than \(max\_doc\_occurrence\) documents. We used coherence to assess which models are better than others.

Based on the parameter search, for LDA, we discard all the tokens that occur in less than 20 documents and the tokens that happen in more than 30\% of all the documents. We also set the number of topics to 10.

For LSI, using the results from parameter search, we set number of topics to 15, \(min\_doc\_occurrence=20\) and \(max\_doc\_occurrence=333181\) which is 30\% of all documents. Table. \ref{tab:hyperparameters}, shows the result of hyperparameters on some of our search experiments.

               
                




\begin{figure}[bt!]
        \centering
        \begin{subfigure}[b]{0.475\textwidth}
            \centering
            \includegraphics[width=\textwidth]{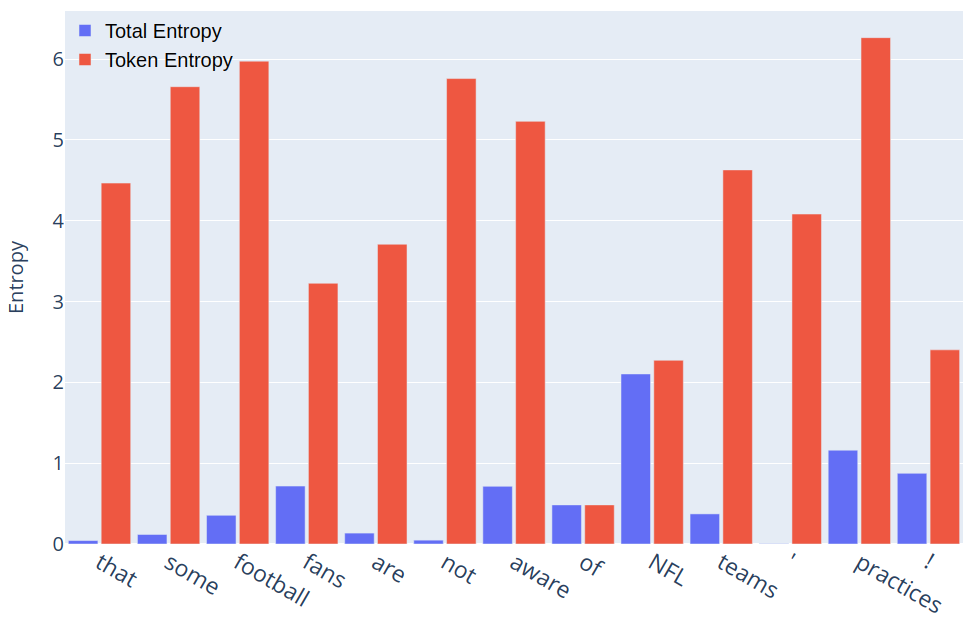}
            \caption[]%
            {{\small Entropy TLG+LSI}}    
            \label{fig:entropy_tlg_lsi}
        \end{subfigure}
        \hfill
        \begin{subfigure}[b]{0.475\textwidth}  
            \centering 
            \includegraphics[width=\textwidth]{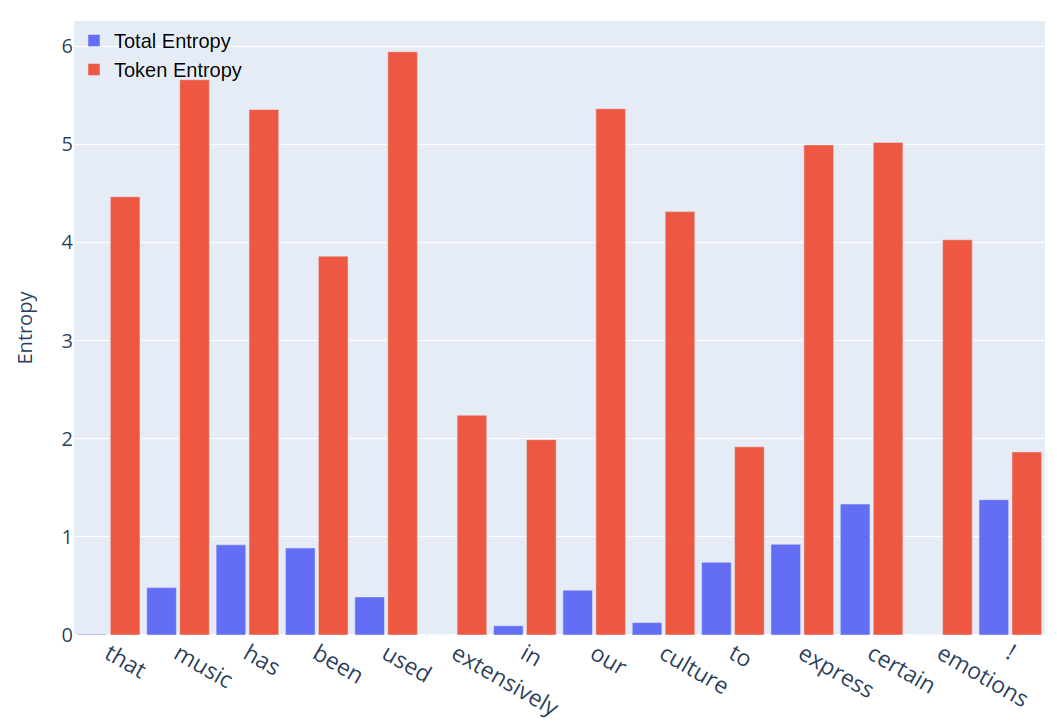}
            \caption[]%
            {{\small  Entropy TLG+LDA}}    
            \label{fig:entropy_tlg_lda}
        \end{subfigure}
        \vskip\baselineskip
        \begin{subfigure}[b]{0.475\textwidth}   
            \centering 
            \includegraphics[width=\textwidth]{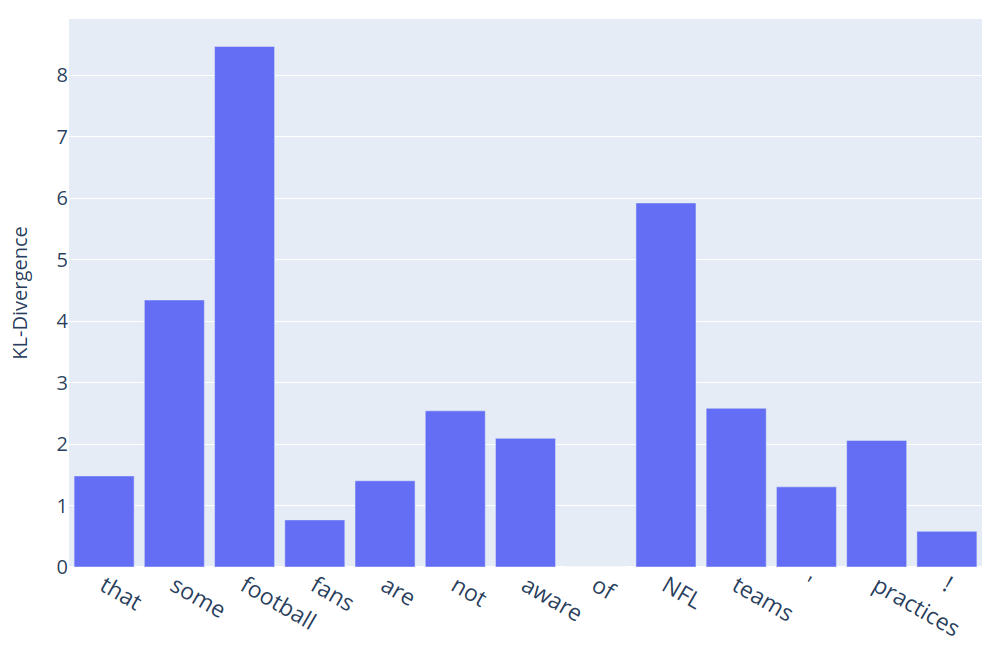}
            \caption[]%
            {{\small KL-Divergence TLG+LSI}}    
            \label{fig:kl_tlg_lsi}
        \end{subfigure}
        \quad
        \begin{subfigure}[b]{0.475\textwidth}   
            \centering 
            \includegraphics[width=\textwidth]{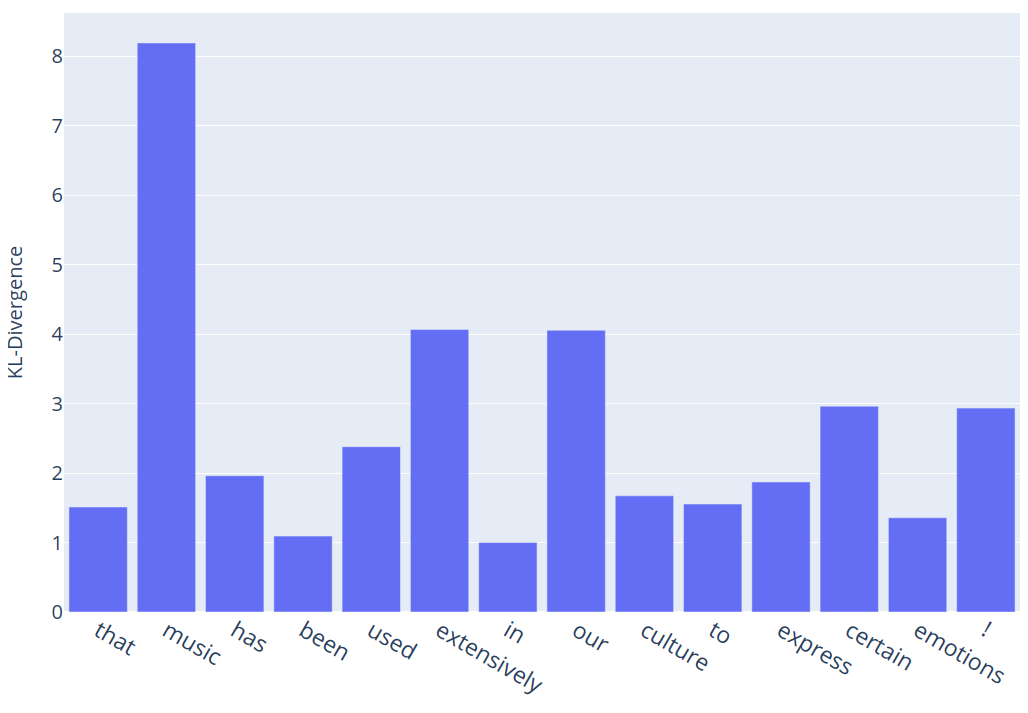}
            \caption[]%
            {{\small KL-Divergence TLG+LDA}}    
            \label{fig:kl_tlg_lda}
        \end{subfigure}
        \caption[ The average and standard deviation of critical parameters ]
        {\small Comparison between the Entropy and KL divergence of TLG with different topic modeling and base GPT-2. Entropy and KL divergence show TLG probabilities (blue) are smaller and, the model is less certain choosing the next token. KL-divergence shows how TLG deviates from the base model on topic tokens} 
        \label{fig:entkldiv}
\end{figure}

\section{Discussion}
\label{sec:discussion}
In this section, we focus on the topical language generation mechanism and how it modifies the probability distribution of the base model. The language generation is the task of generating the next token conditioned on the previously generated tokens. The probability distribution of the next token in the base language models are more flat in some token positions and more peaked at some other token positions. For example, given the prompt of ``The issue is that'' there are plenty of possible next tokens compared to the next token of a prompt like ``It is focused'' which is almost always ``on''. This property of language models gives us the flexibility to meddle in the generation process and steer it towards desired tokens when the probability distribution is flatter.

The concept of flat or peaked distribution can be easily measured in terms of the entropy of the distribution. In Figures \ref{fig:entropy_tlg_lsi} and \ref{fig:entropy_tlg_lda} we compare the entropy of the base model (token entropy) with the posterior probability distribution from Equation \ref{eq:20} as the total entropy. Higher entropy for the base model in one position is a sign of its capability to sample from a large set of potential tokens with almost equal probabilities but in our conditional language modeling, we want to restrict that set to a smaller set that conforms with the chosen topic. Therefore, in almost all cases, the entropy of the TLG model drops significantly compared to the base model. We can observe the differences are larger for the tokens that represent the topic (like teams, football, culture and, music) and smaller for function tokens (like stop words that do not play any role in different topics).

\begin{figure}[bt!]
        \centering
        \begin{subfigure}[b]{0.475\textwidth}
            \centering
            \includegraphics[width=\textwidth]{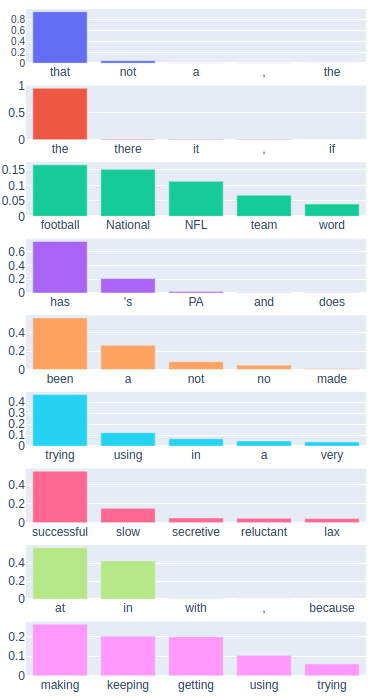}
            \caption[]%
            {{\small Softmax TLG+LSI}}    
            \label{fig:softmax}
        \end{subfigure}
        \hfill
        \begin{subfigure}[b]{0.475\textwidth}  
            \centering 
            \includegraphics[width=\textwidth]{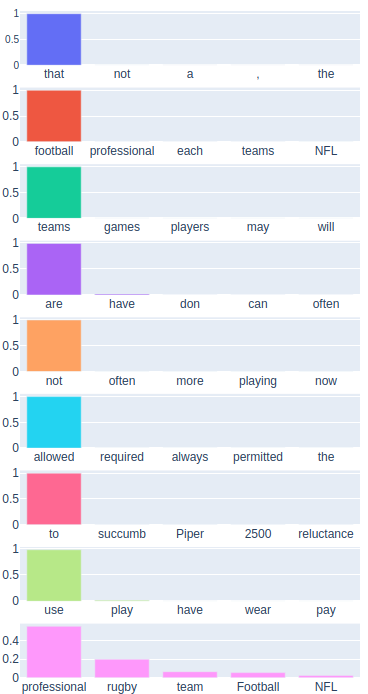}
            \caption[]%
            {{\small  Sparsemax TLG+LDA}}    
            \label{fig:sparsemax}
        \end{subfigure}

        \caption[ ]
        {\small Comparison between the probability of top-5 tokens in softmax and sparsemax: Both functions have candidates that are compatible with topic football. The sparsemax puts less probability on alternatives and that makes it more robust in text generation compared to softmax that always has non zero probability for all tokens in the vocabulary} 
        \label{fig:top5}
\end{figure}

In Figures \ref{fig:kl_tlg_lsi} and \ref{fig:kl_tlg_lda}, the same can be observed for the KL-divergence between the total probability and token probability. In other words, we measure the KL-divergence between the posterior and prior distributions which is the mathematical definition of surprise (\cite{baldi2010bits}):

\begin{equation}
\label{eq:37}
    \operatorname{Surprise}(x_i, t_j |x_{<i})=\operatorname{KL}(P(x_i|x_{<i}, t_j)||P(x_i|x_{<i}))=\sum_{x_i}P(x_i|x_{<i}, t_j)\operatorname{log}(\frac{P(x_i|x_{<i}, t_j)}{P(x_i|x_{x<i})})
\end{equation}

\begin{figure}[bt!]
        \centering
        \begin{subfigure}[b]{0.475\textwidth}
            \centering
            \includegraphics[width=\textwidth]{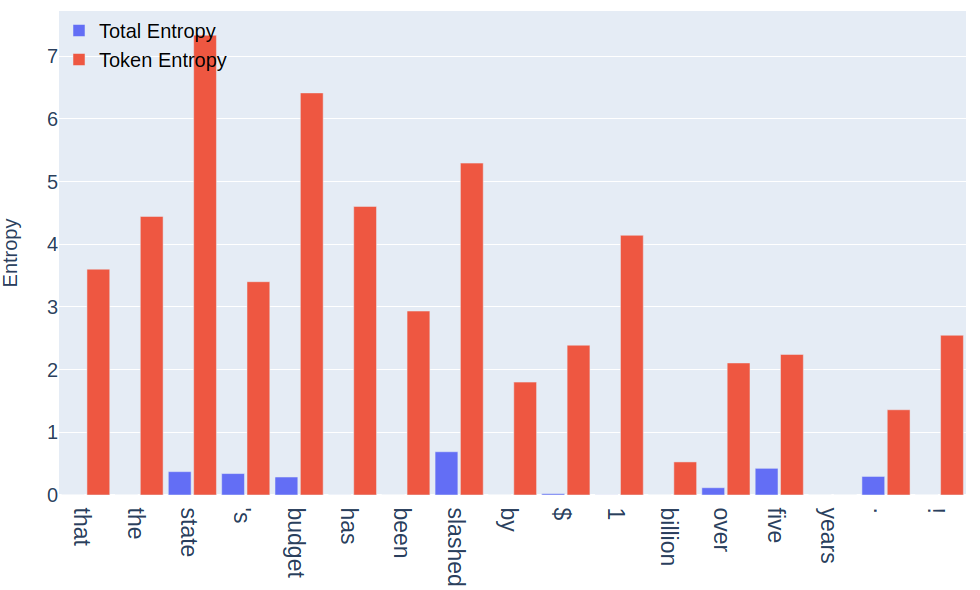}
            \caption[]%
            {{\small Entropy TLG+Softmax}}    
            \label{fig:entropy_tlg_softmax}
        \end{subfigure}
        \hfill
        \begin{subfigure}[b]{0.475\textwidth}  
            \centering 
            \includegraphics[width=\textwidth]{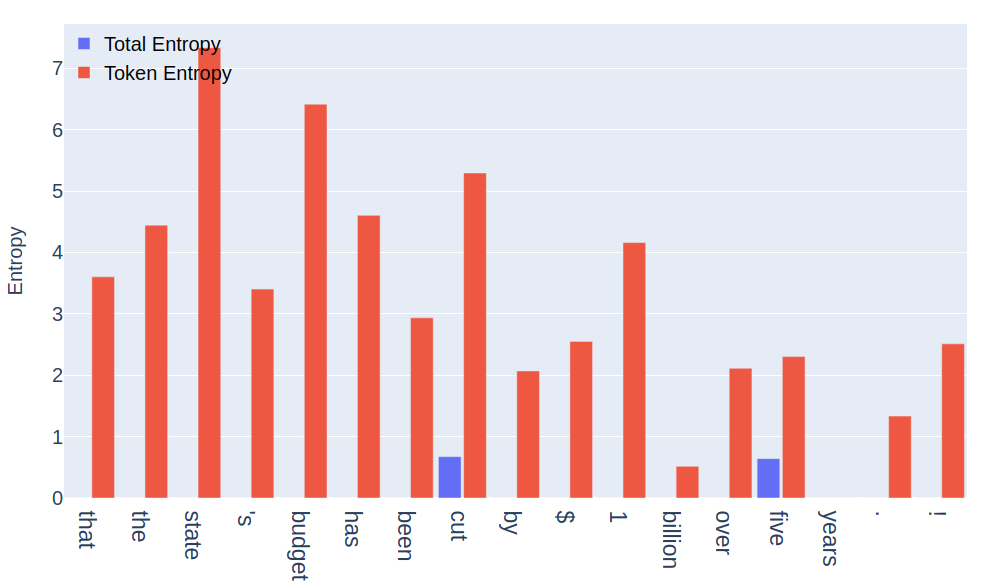}
            \caption[]%
            {{\small  Entropy TLG+Sparsemax}}    
            \label{fig:entropy_tlg_sparsemax}
        \end{subfigure}
        \vskip\baselineskip
        \begin{subfigure}[b]{0.475\textwidth}   
            \centering 
            \includegraphics[width=\textwidth]{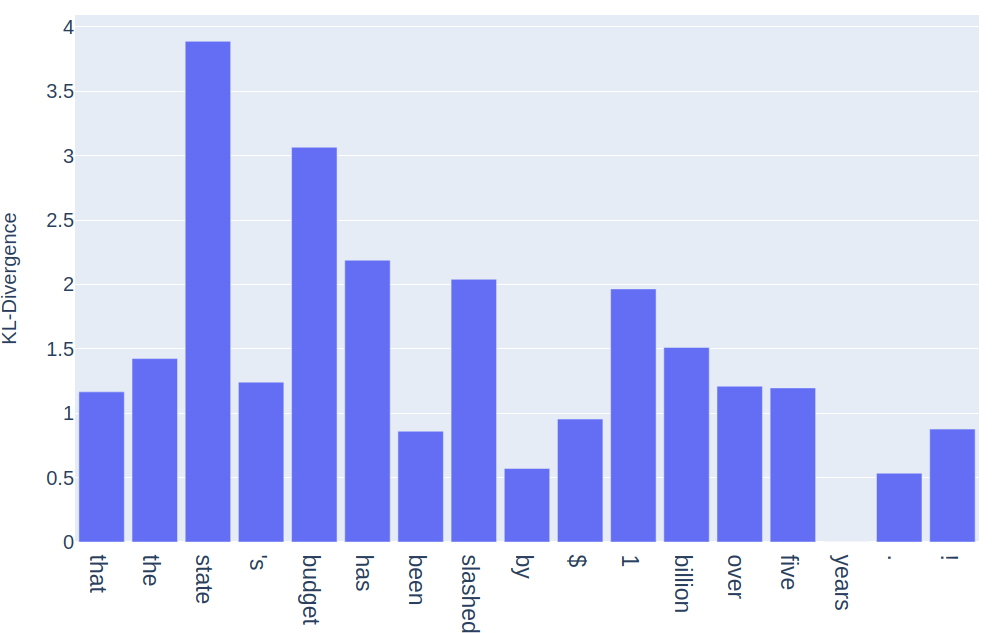}
            \caption[]%
            {{\small KL-Divergence TLG+Softmax}}    
            \label{fig:kl_tlg_softmax}
        \end{subfigure}
        \quad
        \begin{subfigure}[b]{0.475\textwidth}   
            \centering 
            \includegraphics[width=\textwidth]{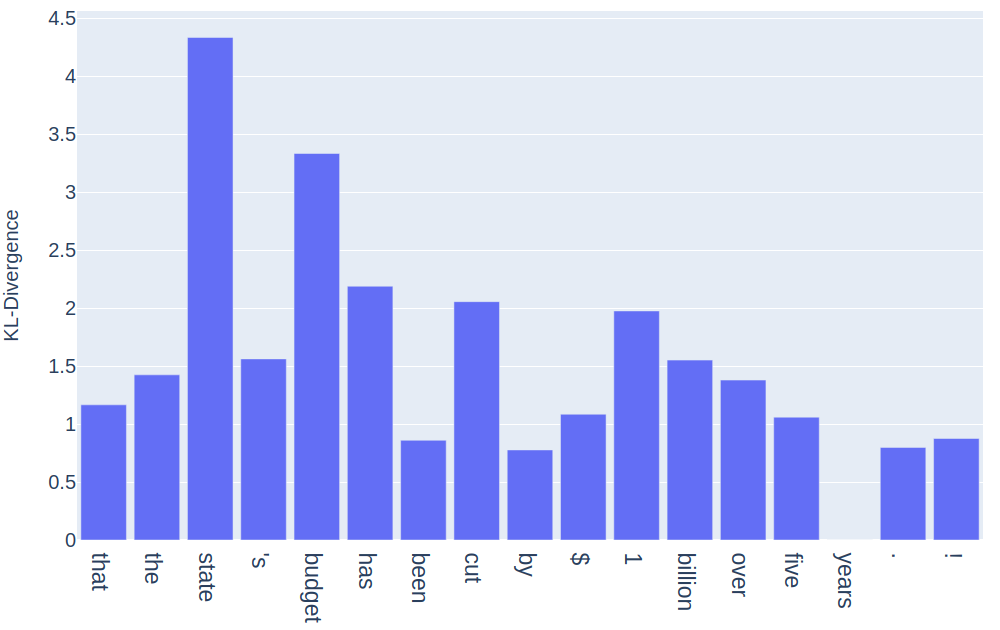}
            \caption[]%
            {{\small KL-Divergence TLG+Sparsemax}}    
            \label{fig:kl_tlg_sparsemax}
        \end{subfigure}
        \caption[  ]
        {\small Comparison between the Entropy and KL divergence of TLG with different activation functions. Entropies of softmax function are more uncertain compared to sparsemax. The KL-divergence between the TLG model and the base model is also wider when sparsemax function has been used} 
        \label{fig:softmaxsparsemax}
\end{figure}

Surprise can also be described as the difference between the cross entropy between TLG and base model and the entropy of TLG:
\begin{equation}
\label{eq:38}
\operatorname{Surprise}(x_i, t_j |x_{<i}) = \operatorname{H}(P(x_i|x_{<i}, t_j)|P(x_i|x_{<i})) - \operatorname{H}(P(x_i|x_{<i}, t_j))
\end{equation}

In this definition, a topic \(t_j\) has no surprise, or new information if it leaves the base language model unaffected. For example, in Figure \ref{fig:kl_tlg_lsi} the token generation that leads to ``of'' is unaffected by the topic of ``football''. On the other hand, if the topic brings new information, the language model will be altered by it. For example, in Figure \ref{fig:kl_tlg_lda} the token generation that leads to the token ``music'' is affected by the chosen topic which is \textit{culture}.

Another interesting observation is how the prior distribution that was extracted from topic modeling forces the language model to choose the topical tokens. The top-5 most likely tokens in a generation process are depicted in Figure \ref{fig:top5}. For the topic of \textit{Football}, the top-5 candidate tokens chosen by the model are compatible with the chosen topic. Both softmax and sparsemax get to choose the relevant candidates for the generation but the softmax function is smoother and as mentioned in Section \ref{sec:controllable_generation_methods} has non-zero probabilities for all tokens that can potentially go off on a tangent by choosing tokens outside the desired tokens of the prior probability. However, the sparsemax function puts less probability, and in most cases even zero probability on out-of-topic tokens. This makes the sparsemax function more robust in topical generation than softmax. Even though sparsemax usually chooses a very small set of candidate tokens, our experiments have shown that it does not affect the overall fluency of the text generation. Table. \ref{tab:comparison} shows that this has been achieved without any detrimental effect on coherence or repetition. 

The difference between the effect of sparsemax versus softmax function also can be observed by carefully choosing two very close examples. As mentioned above, the sparsemax function has less uncertainty in choosing the next token and it can be seen in Figure \ref{fig:entropy_tlg_sparsemax} when the model has to choose the word ``state'' in the topic of \textit{politics}. This behavior also results in more divergence from the base model because sparsemax comes with more on-topic and certain choices. Figures \ref{fig:kl_tlg_softmax} and \ref{fig:kl_tlg_sparsemax}, show that even though the shape of KL-divergence in both cases are almost the same, with sparsemax the difference is larger.

Since we use Byte Pair Encoding (BPE) tokenized inputs for our topic modeling, it is usual to see at least one topic with token sets for topics that are not a complete word. For example, in Table \ref{tab:bad_examples} the first and second rows show tokens: ``qu'', ``\&'', ``Earth'', ``rs'', ``her'', ``ld'', ``ld'', ``rd'', ``she'', ``we'', most of which are not complete words. The same phenomena has been observed for both LDA and LSI. The resulting generated text from these topics is not fluent. It is also full of acronyms, links, and even non-English characters has been produced. The second row shows a generated text that is more fluent but still does not make sense and is full of glitches. Another case in topic models is when the topic is vague and the words in it do not seem to belong to one topic. For example, the third row shows a topic with tokens: ``comedy'', ``social'', ``comic'', ``public'', ``Greek'', ``students'', ``company'' that does not conform to one topic. In this case, the generated text degenerates at the end. In our experiments, we observed that this problem is less likely to happen with the LSI model compared to LDA for topic models. For example, the last row shows one case of a topic that does not have complete words in it except the last two. Even in this case, the quality of the generated text is better than others and its relevance to the last two tokens ``medium'' and ``television'' has been represented more accurately.

\begin{table}[]
\caption{Samples from bad examples generated by TLG model. The first, second and, last rows show examples of “non-word” topic top tokens for the task of text generation and the third row shows an example of a vague topic}
\label{tab:bad_examples}
\begin{tabular}{ccl}
\hline
Topic top tokens                                                                                                      & Model & \multicolumn{1}{c}{Generated text}                                                                                                                        \\ \hline
\begin{tabular}[c]{@{}c@{}}'qu', '\&', 'Earth',\\  'rs', 'her', 'ld', \\ 'rd', 'she', 'we'\end{tabular}              & LDA   & \begin{tabular}[c]{@{}l@{}}The issue is that the company has been accused of using "the social media platform \\ to promote its products and services.".@SocietyInTupelo public Facebook page,\\  which was used by @ianeste for a few months now— https://t:co/Yq6J5Xj7x4\\  — Greekpic.@Societates political party in \#IPL\_in \\ February 2016 Google+ view All photos © 2014 any user at largest\end{tabular}                                                                                                                                                                                                                                                                                          \\ \hline
\begin{tabular}[c]{@{}c@{}}'qu', '\&', 'Earth',\\  'rs', 'her', 'ld', \\ 'rd', 'she', 'we'\end{tabular}              & LDA   & \begin{tabular}[c]{@{}l@{}}The issue is that the new version of Earth's magnetic field has been so weak, \\ it can't be used as a magnet because there aren"quasitant magnets." The problem?\\  It doesn. "It takes about 10 percent less energy to create an electric current than \\ would normally occur in nature," she says {[}1{]} But this isn't enough if we're going like \\ herrsy \& Healy just now where you have all those super good things happening \\ at once and then!\end{tabular}                                                                                                                                                                                                       \\ \hline
\begin{tabular}[c]{@{}c@{}}'comedy', 'social', \\ 'comic', 'public',\\ 'Greek', 'students',\\ 'company'\end{tabular} & LDA   & \begin{tabular}[c]{@{}l@{}}He goes to public meetings and he's a very good speaker. He has the ability\\ —the capacity —to make people feel comfortable with him without being\\  too judgmental." (Photo courtesy Facebook.) It was created by an anonymous user\\  who posted about his experience at comedy clubs in New York City during one \\ comic strip called 4 billion dollars worth of jokes from comedians like Bill Maher\\  any day later when it became clear they were  not going anywhere because their\\ audience hadalready seen them perform for free online before getting paid \$10m\\ per episode far: http://idianateseestinireusesfromandropollimeloemokesizedramaI'm\end{tabular} \\ \hline
\begin{tabular}[c]{@{}c@{}}'qu', '\&', 'o', \\ 'rs', 'ld', 'rd',\\ 'medium', \\ 'television'\end{tabular}            & LSI   & \begin{tabular}[c]{@{}l@{}}The issue is that the medium used here does not have telecommunication \\ capabilities. In other words, it's a very small area where you can't really\\ use your phone to communicate with people in real time when they're on\\ television." I asked he about what would happen if we had an Internet service\\ provider like Comcast or Verizon and said there were no plans for such services at all\end{tabular}                                                                                                                                                                                                                                                             \\ \hline
\end{tabular}
\end{table}

\section{Conclusion}
In this paper, we introduced Topical Language Generation using transformers. Our approach lays out a very simple way to assemble any transformer-based model with a topical language model trained on a corpus of choice to produce high quality controlled text. The Bayesian approach helped us to verify that the controlled language generation is the posterior probability with base LM being likelihood and the topic modeling being prior. We believe our approach can be used as a baseline for any open-ended controlled language generation task that is challenging. 
The generality of our approach can be used as a foundation for any controlled language generation task. For example, we hope to extend this work to include more diverse priors such as sentiment, formality, and style. The choice of topic models also can be extended to other approaches that gather distributional properties of words on a control variable. For example approaches that use word embeddings learned conditioned on topics or other variables are also can be used. 
The ever-increasing power of LMs still needs better decoding techniques that our approach has achieved but more importantly, it opens the door for even more exciting research in the future.

\bibliographystyle{nlelike}
\bibliography{main}

\begin{thebibliography}{}

\bibitem[Baheti et~al., 2018]{baheti2018generating}
{\bf Baheti, A.}, {\bf Ritter, A.}, {\bf Li, J.}, \textbf{and} {\bf Dolan, B.}
  2018.
\newblock Generating more interesting responses in neural conversation models
  with distributional constraints.
\newblock In {\em Proceedings of the 2018 Conference on Empirical Methods in
  Natural Language Processing}, pp. 3970--3980, Brussels, Belgium. Association
  for Computational Linguistics.

\bibitem[Baldi and Itti, 2010]{baldi2010bits}
{\bf Baldi, P.} \textbf{and} {\bf Itti, L.} 2010.
\newblock Of bits and wows: A bayesian theory of surprise with applications to
  attention.
\newblock {\em Neural Networks}, 23(5):649--666.

\bibitem[Blei et~al., 2003]{blei2003latent}
{\bf Blei, D.~M.}, {\bf Ng, A.~Y.}, \textbf{and} {\bf Jordan, M.~I.} 2003.
\newblock Latent dirichlet allocation.
\newblock {\em Journal of machine Learning research}, 3(Jan):993--1022.

\bibitem[Bowman et~al., 2015]{bowman2015generating}
{\bf Bowman, S.~R.}, {\bf Vilnis, L.}, {\bf Vinyals, O.}, {\bf Dai, A.~M.},
  {\bf Jozefowicz, R.}, \textbf{and} {\bf Bengio, S.} 2015.
\newblock Generating sentences from a continuous space.
\newblock {\em SIGNLL Conference on Computational Natural Language Learning
  (CONLL)}.

\bibitem[Brown et~al., 2020]{brown2020language}
{\bf Brown, T.~B.}, {\bf Mann, B.}, {\bf Ryder, N.}, {\bf Subbiah, M.}, {\bf
  Kaplan, J.}, {\bf Dhariwal, P.}, {\bf Neelakantan, A.}, {\bf Shyam, P.}, {\bf
  Sastry, G.}, {\bf Askell, A.}, \textbf{and} {\bf others} 2020.
\newblock Language models are few-shot learners.
\newblock {\em Advances in Neural Information Processing Systems 33 (NeurIPS
  2020)}.

\bibitem[Correia et~al., 2019]{correia2019adaptively}
{\bf Correia, G.~M.}, {\bf Niculae, V.}, \textbf{and} {\bf Martins, A. F.~T.}
  2019.
\newblock Adaptively sparse transformers.
\newblock In {\em Proceedings of the 2019 Conference on Empirical Methods in
  Natural Language Processing and the 9th International Joint Conference on
  Natural Language Processing (EMNLP-IJCNLP)}, pp. 2174--2184, Hong Kong,
  China. Association for Computational Linguistics.

\bibitem[Deerwester et~al., 1990]{deerwester1990indexing}
{\bf Deerwester, S.}, {\bf Dumais, S.~T.}, {\bf Furnas, G.~W.}, {\bf Landauer,
  T.~K.}, \textbf{and} {\bf Harshman, R.} 1990.
\newblock Indexing by latent semantic analysis.
\newblock {\em Journal of the American society for information science},
  41(6):391--407.

\bibitem[Dethlefs and Cuay{\'a}huitl, 2015]{dethlefs2015hierarchical}
{\bf Dethlefs, N.} \textbf{and} {\bf Cuay{\'a}huitl, H.} 2015.
\newblock Hierarchical reinforcement learning for situated natural language
  generation.
\newblock {\em Natural Language Engineering}, 21(3).

\bibitem[Dziri et~al., 2018]{dziri2018augmenting}
{\bf Dziri, N.}, {\bf Kamalloo, E.}, {\bf Mathewson, K.~W.}, \textbf{and} {\bf
  Zaiane, O.} 2018.
\newblock Augmenting neural response generation with context-aware topical
  attention.
\newblock {\em Proceedings of the First Workshop on NLP for Conversational AI}.

\bibitem[Fu et~al., 2018]{fu2017style}
{\bf Fu, Z.}, {\bf Tan, X.}, {\bf Peng, N.}, {\bf Zhao, D.}, \textbf{and} {\bf
  Yan, R.} 2018.
\newblock Style transfer in text: Exploration and evaluation.
\newblock In {\em Proceedings of the AAAI Conference on Artificial
  Intelligence}, volume~32.

\bibitem[Gage, 1994]{gage1994new}
{\bf Gage, P.} 1994.
\newblock A new algorithm for data compression.
\newblock {\em C Users Journal}, 12(2):23--38.

\bibitem[Ghazvininejad et~al., 2017]{ghazvininejad2017hafez}
{\bf Ghazvininejad, M.}, {\bf Shi, X.}, {\bf Priyadarshi, J.}, \textbf{and}
  {\bf Knight, K.} 2017.
\newblock Hafez: an interactive poetry generation system.
\newblock In {\em Proceedings of ACL 2017, System Demonstrations}, pp. 43--48.

\bibitem[Goodfellow, 2016]{goodfellow2016nips}
{\bf Goodfellow, I.} 2016.
\newblock Nips 2016 tutorial: Generative adversarial networks.
\newblock {\em arXiv preprint arXiv:1701.00160}.

\bibitem[Gopalakrishnan et~al., 2019]{gopalakrishnan2019topical}
{\bf Gopalakrishnan, K.}, {\bf Hedayatnia, B.}, {\bf Chen, Q.}, {\bf Gottardi,
  A.}, {\bf Kwatra, S.}, {\bf Venkatesh, A.}, {\bf Gabriel, R.}, {\bf
  Hakkani-T{\"u}r, D.}, \textbf{and} {\bf AI, A.~A.} 2019.
\newblock Topical-chat: Towards knowledge-grounded open-domain conversations.
\newblock {\em Proc. Interspeech 2019}, pp. 1891--1895.

\bibitem[Guo et~al., 2018]{guo2018long}
{\bf Guo, J.}, {\bf Lu, S.}, {\bf Cai, H.}, {\bf Zhang, W.}, {\bf Yu, Y.},
  \textbf{and} {\bf Wang, J.} 2018.
\newblock Long text generation via adversarial training with leaked
  information.
\newblock In {\em Thirty-Second AAAI Conference on Artificial Intelligence}.

\bibitem[Halko et~al., 2011]{halko2011finding}
{\bf Halko, N.}, {\bf Martinsson, P.-G.}, \textbf{and} {\bf Tropp, J.~A.} 2011.
\newblock Finding structure with randomness: Probabilistic algorithms for
  constructing approximate matrix decompositions.
\newblock {\em SIAM review}, 53(2):217--288.

\bibitem[Hoffman et~al., 2010]{hoffman2010online}
{\bf Hoffman, M.}, {\bf Bach, F.~R.}, \textbf{and} {\bf Blei, D.~M.} 2010.
\newblock Online learning for latent dirichlet allocation.
\newblock In {\em advances in neural information processing systems}, pp.
  856--864.

\bibitem[Holtzman et~al., 2020]{holtzman2019curious}
{\bf Holtzman, A.}, {\bf Buys, J.}, {\bf Du, L.}, {\bf Forbes, M.},
  \textbf{and} {\bf Choi, Y.} 2020.
\newblock The curious case of neural text degeneration.
\newblock In {\em International Conference on Learning Representations}.

\bibitem[Holtzman et~al., 2018]{holtzman2018learning}
{\bf Holtzman, A.}, {\bf Buys, J.}, {\bf Forbes, M.}, {\bf Bosselut, A.}, {\bf
  Golub, D.}, \textbf{and} {\bf Choi, Y.} 2018.
\newblock Learning to write with cooperative discriminators.
\newblock {\em In Proceedings of ACL}.

\bibitem[Hu et~al., 2017]{hu2017toward}
{\bf Hu, Z.}, {\bf Yang, Z.}, {\bf Liang, X.}, {\bf Salakhutdinov, R.},
  \textbf{and} {\bf Xing, E.~P.} 2017.
\newblock Toward controlled generation of text.
\newblock In {\em Proceedings of the 34th International Conference on Machine
  Learning-Volume 70}, pp. 1587--1596. JMLR. org.

\bibitem[Huang, 2005]{huang2005maximum}
{\bf Huang, J.} 2005.
\newblock Maximum likelihood estimation of dirichlet distribution parameters.
\newblock {\em CMU Technique Report}.

\bibitem[Kannan et~al., 2016]{kannan2016smart}
{\bf Kannan, A.}, {\bf Kurach, K.}, {\bf Ravi, S.}, {\bf Kaufmann, T.}, {\bf
  Tomkins, A.}, {\bf Miklos, B.}, {\bf Corrado, G.}, {\bf Lukacs, L.}, {\bf
  Ganea, M.}, {\bf Young, P.}, \textbf{and} {\bf others} 2016.
\newblock Smart reply: Automated response suggestion for email.
\newblock In {\em Proceedings of the 22nd ACM SIGKDD International Conference
  on Knowledge Discovery and Data Mining}, pp. 955--964.

\bibitem[Keskar et~al., 2019]{keskar2019ctrl}
{\bf Keskar, N.~S.}, {\bf McCann, B.}, {\bf Varshney, L.~R.}, {\bf Xiong, C.},
  \textbf{and} {\bf Socher, R.} 2019.
\newblock Ctrl: A conditional transformer language model for controllable
  generation.
\newblock {\em arXiv preprint arXiv:1909.05858}.

\bibitem[Lau et~al., 2017]{lau2017topically}
{\bf Lau, J.~H.}, {\bf Baldwin, T.}, \textbf{and} {\bf Cohn, T.} 2017.
\newblock Topically driven neural language model.
\newblock In {\em Proceedings of the 55th Annual Meeting of the Association for
  Computational Linguistics (Volume 1: Long Papers)}, pp. 355--365, Vancouver,
  Canada. Association for Computational Linguistics.

\bibitem[Li et~al., 2018]{li2018delete}
{\bf Li, J.}, {\bf Jia, R.}, {\bf He, H.}, \textbf{and} {\bf Liang, P.} 2018.
\newblock Delete, retrieve, generate: a simple approach to sentiment and style
  transfer.
\newblock In {\em Proceedings of the 2018 Conference of the North {A}merican
  Chapter of the Association for Computational Linguistics: Human Language
  Technologies, Volume 1 (Long Papers)}, pp. 1865--1874, New Orleans,
  Louisiana. Association for Computational Linguistics.

\bibitem[Li et~al., 2020]{li2019don}
{\bf Li, M.}, {\bf Roller, S.}, {\bf Kulikov, I.}, {\bf Welleck, S.}, {\bf
  Boureau, Y.-L.}, {\bf Cho, K.}, \textbf{and} {\bf Weston, J.} 2020.
\newblock Don{'}t say that! making inconsistent dialogue unlikely with
  unlikelihood training.
\newblock In {\em Proceedings of the 58th Annual Meeting of the Association for
  Computational Linguistics}, pp. 4715--4728, Online. Association for
  Computational Linguistics.

\bibitem[Malandrakis et~al., 2019a]{malandrakis2019controlled}
{\bf Malandrakis, N.}, {\bf Shen, M.}, {\bf Goyal, A.}, {\bf Gao, S.}, {\bf
  Sethi, A.}, \textbf{and} {\bf Metallinou, A.} 2019a.
\newblock Controlled text generation for data augmentation in intelligent
  artificial agents.
\newblock {\em arXiv preprint arXiv:1910.03487}.

\bibitem[Malandrakis et~al., 2019b]{dathathri2019plug}
{\bf Malandrakis, N.}, {\bf Shen, M.}, {\bf Goyal, A.}, {\bf Gao, S.}, {\bf
  Sethi, A.}, \textbf{and} {\bf Metallinou, A.} 2019b.
\newblock Controlled text generation for data augmentation in intelligent
  artificial agents.
\newblock In {\em Proceedings of the 3rd Workshop on Neural Generation and
  Translation}, pp. 90--98, Hong Kong. Association for Computational
  Linguistics.

\bibitem[Martins and Astudillo, 2016]{martins2016softmax}
{\bf Martins, A.} \textbf{and} {\bf Astudillo, R.} 2016.
\newblock From softmax to sparsemax: A sparse model of attention and
  multi-label classification.
\newblock In {\em International Conference on Machine Learning}, pp.
  1614--1623.

\bibitem[Mikolov et~al., 2013]{mikolov2013distributed}
{\bf Mikolov, T.}, {\bf Sutskever, I.}, {\bf Chen, K.}, {\bf Corrado, G.~S.},
  \textbf{and} {\bf Dean, J.} 2013.
\newblock Distributed representations of words and phrases and their
  compositionality.
\newblock In {\em Advances in neural information processing systems}, pp.
  3111--3119.

\bibitem[Mueller et~al., 2017]{mueller2017sequence}
{\bf Mueller, J.}, {\bf Gifford, D.}, \textbf{and} {\bf Jaakkola, T.} 2017.
\newblock Sequence to better sequence: continuous revision of combinatorial
  structures.
\newblock In {\em International Conference on Machine Learning}, pp.
  2536--2544.

\bibitem[Petroni et~al., 2019]{petroni2019language}
{\bf Petroni, F.}, {\bf Rockt{\"a}schel, T.}, {\bf Lewis, P.}, {\bf Bakhtin,
  A.}, {\bf Wu, Y.}, {\bf Miller, A.~H.}, \textbf{and} {\bf Riedel, S.} 2019.
\newblock Language models as knowledge bases?
\newblock {\em EMNLP}.

\bibitem[Prabhumoye et~al., 2018]{prabhumoye2018style}
{\bf Prabhumoye, S.}, {\bf Tsvetkov, Y.}, {\bf Salakhutdinov, R.}, \textbf{and}
  {\bf Black, A.~W.} 2018.
\newblock Style transfer through back-translation.
\newblock {\em ACL}.

\bibitem[Radford et~al., 2019]{radford2019language}
{\bf Radford, A.}, {\bf Wu, J.}, {\bf Child, R.}, {\bf Luan, D.}, {\bf Amodei,
  D.}, \textbf{and} {\bf Sutskever, I.} 2019.
\newblock Language models are unsupervised multitask learners.
\newblock {\em OpenAI Blog}, 1(8):9.

\bibitem[Reimers and Gurevych, 2019]{reimers-2019-sentence-bert}
{\bf Reimers, N.} \textbf{and} {\bf Gurevych, I.} 2019.
\newblock Sentence-bert: Sentence embeddings using siamese bert-networks.
\newblock In {\em Proceedings of the 2019 Conference on Empirical Methods in
  Natural Language Processing}. Association for Computational Linguistics.

\bibitem[R{\"o}der et~al., 2015]{roder2015exploring}
{\bf R{\"o}der, M.}, {\bf Both, A.}, \textbf{and} {\bf Hinneburg, A.} 2015.
\newblock Exploring the space of topic coherence measures.
\newblock In {\em Proceedings of the eighth ACM international conference on Web
  search and data mining}, pp. 399--408.

\bibitem[See et~al., 2019]{see2019makes}
{\bf See, A.}, {\bf Roller, S.}, {\bf Kiela, D.}, \textbf{and} {\bf Weston, J.}
  2019.
\newblock What makes a good conversation? how controllable attributes affect
  human judgments.
\newblock In {\em Proceedings of the 2019 Conference of the North {A}merican
  Chapter of the Association for Computational Linguistics: Human Language
  Technologies, Volume 1 (Long and Short Papers)}, pp. 1702--1723, Minneapolis,
  Minnesota. Association for Computational Linguistics.

\bibitem[Singh and Palod, 2018]{singh2018sentiment}
{\bf Singh, A.} \textbf{and} {\bf Palod, R.} 2018.
\newblock Sentiment transfer using seq2seq adversarial autoencoders.
\newblock {\em arXiv preprint arXiv:1804.04003}.

\bibitem[Stahlberg et~al., 2018]{stahlberg2018simple}
{\bf Stahlberg, F.}, {\bf Cross, J.}, \textbf{and} {\bf Stoyanov, V.} 2018.
\newblock Simple fusion: Return of the language model.
\newblock {\em WMT18}.

\bibitem[Tsallis, 1988]{tsallis1988possible}
{\bf Tsallis, C.} 1988.
\newblock Possible generalization of boltzmann-gibbs statistics.
\newblock {\em Journal of statistical physics}, 52(1-2):479--487.

\bibitem[Vaswani et~al., 2017]{vaswani2017attention}
{\bf Vaswani, A.}, {\bf Shazeer, N.}, {\bf Parmar, N.}, {\bf Uszkoreit, J.},
  {\bf Jones, L.}, {\bf Gomez, A.~N.}, {\bf Kaiser, {\L}.}, \textbf{and} {\bf
  Polosukhin, I.} 2017.
\newblock Attention is all you need.
\newblock In {\em Advances in neural information processing systems}, pp.
  5998--6008.

\bibitem[Welleck et~al., 2020]{welleck2019neural}
{\bf Welleck, S.}, {\bf Kulikov, I.}, {\bf Roller, S.}, {\bf Dinan, E.}, {\bf
  Cho, K.}, \textbf{and} {\bf Weston, J.} 2020.
\newblock Neural text generation with unlikelihood training.
\newblock In {\em International Conference on Learning Representations}.

\bibitem[Xing et~al., 2017]{xing2017topic}
{\bf Xing, C.}, {\bf Wu, W.}, {\bf Wu, Y.}, {\bf Liu, J.}, {\bf Huang, Y.},
  {\bf Zhou, M.}, \textbf{and} {\bf Ma, W.-Y.} 2017.
\newblock Topic aware neural response generation.
\newblock In {\em Thirty-First AAAI Conference on Artificial Intelligence}.

\bibitem[Xu et~al., 2018]{xu2018unpaired}
{\bf Xu, J.}, {\bf Sun, X.}, {\bf Zeng, Q.}, {\bf Ren, X.}, {\bf Zhang, X.},
  {\bf Wang, H.}, \textbf{and} {\bf Li, W.} 2018.
\newblock Unpaired sentiment-to-sentiment translation: A cycled reinforcement
  learning approach.
\newblock {\em ACL}.

\bibitem[Yu et~al., 2017]{yu2017seqgan}
{\bf Yu, L.}, {\bf Zhang, W.}, {\bf Wang, J.}, \textbf{and} {\bf Yu, Y.} 2017.
\newblock Seqgan: Sequence generative adversarial nets with policy gradient.
\newblock In {\em Thirty-First AAAI Conference on Artificial Intelligence}.

\bibitem[Zhang et~al., 2018]{zhang2018shaped}
{\bf Zhang, Y.}, {\bf Ding, N.}, \textbf{and} {\bf Soricut, R.} 2018.
\newblock {SHAPED}: Shared-private encoder-decoder for text style adaptation.
\newblock In {\em Proceedings of the 2018 Conference of the North {A}merican
  Chapter of the Association for Computational Linguistics: Human Language
  Technologies, Volume 1 (Long Papers)}, pp. 1528--1538, New Orleans,
  Louisiana. Association for Computational Linguistics.

\bibitem[Zhao et~al., 2018]{zhao2018language}
{\bf Zhao, Y.}, {\bf Bi, V.~W.}, {\bf Cai, D.}, {\bf Liu, X.}, {\bf Tu, K.},
  \textbf{and} {\bf Shi, S.} 2018.
\newblock Language style transfer from non-parallel text with arbitrary styles.
\newblock {\em International Conference on Learning Representations}.
\newblock rejected.

\end{thebibliography}

\label{lastpage}

\end{document}